\documentclass{article}

\PassOptionsToPackage{numbers,compress}{natbib}

    \usepackage[preprint]{neurips_2024}

\usepackage[utf8]{inputenc} 
\usepackage[T1]{fontenc}    
\usepackage{hyperref}       
\usepackage{url}            
\usepackage{booktabs}       
\usepackage{amsfonts}       
\usepackage{nicefrac}       
\usepackage{microtype}      
\usepackage{xcolor}         

\usepackage{graphicx}
\usepackage{booktabs}
\usepackage{indentfirst}
\usepackage{multirow}
\usepackage{hhline}
\usepackage{verbatim}
\usepackage{subcaption}
\usepackage{tabularx}
\usepackage{amsmath}

\DeclareMathOperator*{\argmax}{argmax}

\makeatletter
    \def\@fnsymbol#1{\ensuremath{\ifcase#1\or \ddagger\or \dagger\or
       \mathsection\or \mathparagraph\or \|\or **\or \dagger\dagger
       \or \ddagger\ddagger \else\@ctrerr\fi}}
\makeatother

\title{Balancing Efficiency and Quality: MoEISR for Arbitrary-Scale Image Super-Resolution}

\author{%
  Young Jae Oh$^1$, Jihun Kim$^1$, Jihoon Nam, Tae Hyun Kim\thanks{Corresponding author} \\
  Hanyang University, Seoul, South Korea \\
  \texttt{naohjae, jihunkim, namghoon, taehyunkim@hanyang.ac.kr}
}

\usepackage{xspace}
\makeatletter
\def\@onedot{\ifx\@let@token.\else.\null\fi\xspace}
\makeatother
\newcommand{\eg}{e.g\onedot}

\begin{document}

\maketitle

\begin{abstract}
   Arbitrary-scale image super-resolution employing implicit neural functions has gained significant attention lately due to its capability to upscale images across diverse scales utilizing only a single model.
   Nevertheless, these methodologies have imposed substantial computational demands as they involve querying every target pixel to a single resource-intensive decoder.
   In this paper, we introduce a novel and efficient framework, the Mixture-of-Experts Implicit Super-Resolution (MoEISR), which enables super-resolution at arbitrary scales with significantly increased computational efficiency without sacrificing reconstruction quality.
   MoEISR dynamically allocates the most suitable decoding expert to each pixel using a lightweight mapper module, allowing experts with varying capacities to reconstruct pixels across regions with diverse complexities. 
   Our experiments demonstrate that MoEISR successfully reduces significant amount of floating point operations (FLOPs) while delivering comparable or superior peak signal-to-noise ratio (PSNR).
\end{abstract}

\section{Introduction}
\label{sec:intro}
Single image super-resolution (SISR), which aims to reconstruct a high-resolution (HR) image from a low-resolution (LR) image, stands as a foundational task in computer vision given its capacity to serve the input for various other vision applications. While the majority of recent super-resolution (SR) techniques utilize PixelShuffle~\cite{Authors27} based on the convolutional layers as the decoding function~\cite{Authors1, Authors15, Authors16, Authors17, authors34, authors35}, the approach exhibits a notable limitation in that it can only upscale images with a predetermined scale. 
To address this constraint, arbitrary-scale SR methods~\cite{Authors13, Authors14, authors31, authors32, authors33, authors36, authors37, authors38} have been introduced, most of them incorporating a multi-layer perceptron (MLP) as the decoder.
\begin{figure}
    \centering
    \includegraphics[width=0.8\linewidth]{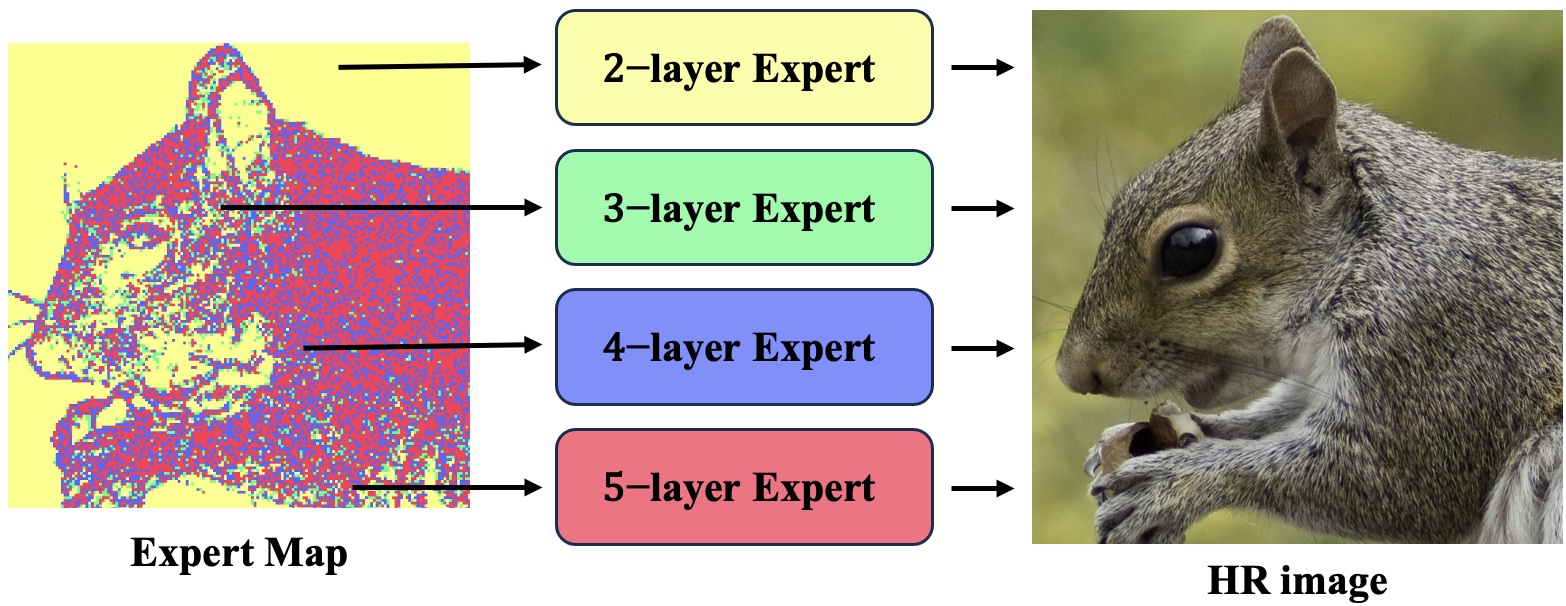}
    \caption{\textbf{Expert map from the mapper.} Yellow, green, blue and red pixels in the expert map denote varying levels of reconstruction complexity and their respective associated experts.}
    \label{fig:1}
    \vspace{-3mm}
\end{figure}
These methods can handle SR at various target scales, which is advantageous since only a single model is required for upscaling images to various resolutions. Nonetheless, these methods place a substantial computational burden, as each pixel necessitates processing by a single resource-intensive MLP decoder. 
Consequently, they still struggle to provide the essential efficiency required for practical applications, particularly when dealing with the reconstruction of images on a large scale.
Furthermore, these studies predominantly neglect the computational burden while emphasizing the network's reconstruction capability.

In this paper, we introduce the Mixture-of-Experts Implicit Super-Resolution (MoEISR) for arbitrary-scale SR, achieving a substantial FLOPs reduction in comparison to prevailing arbitrary-scale SR networks, while delivering comparable or even superior PSNR. To produce high-quality HR images with an emphasis on computational efficiency, our approach involves the joint training of several components. This includes an encoder responsible for generating implicit neural representation (INR), a set of capacity-varying MLP decoders (experts) designed to predict the RGB value of each pixel, and a mapper that assigns pixels to the appropriate expert.

The efficacy of MoEISR lies in the mapper, which adeptly assigns experts with varying capacities to each output pixel.
In our endeavor to harness the potential of capacity-varying experts while maintaining high reconstruction quality, 
we integrate the mixture-of-experts (MoE) scheme~\cite{authors44, authors45, authors46, authors51} which allows the router (mapper) to assign capacity-varying experts to each output pixel.

To illustrate the process, an LR image is passed through the encoder to obtain the INR, which is processed by the mapper to generate an expert map assigning suitable experts to each pixel.
The obatined INR and target coordinates are sent to the assigned expert, predicting the RGB value.
The expert map generated by the mapper offers a distinct visualization of the spatial relationships and relative complexity among pixels. 
The expert map depicted in Fig.~\ref{fig:1} offers a critical insight into the efficiency of our model. By allocating more computationally intensive experts  (\eg ~4-layer and 5-layer experts) to regions of the image with higher complexity, such as object boundaries with abrupt RGB value transitions and areas with intricate textures, the model can handle these challenging areas more effectively with decreased computational demand. This targeted allocation ensures that complex regions receive the necessary processing power, optimizing the balance between performance and computational efficiency.
Given its model-agnostic nature, our MoEISR seamlessly integrates into various INR-based arbitrary-scale SR networks. This integration enhances HR reconstruction quality while reducing computational complexity, demonstrating MoEISR's adaptability across diverse arbitrary-scale SR frameworks and contributing to the advancement of image upscaling techniques.
Furthermore, while our experiments demonstrate the effectiveness of our approach in arbitrary-scale SR, our method is also applicable to various research domains utilizing INR, such as 3D modeling.

We summarize our major contributions as follows:
\begin{itemize}
  \item We present a model-agnostic approach for INR-based arbitrary-scale SR networks.
  \item We demonstrate that MoEISR achieves comparable or superior results to existing arbitrary-scale SR networks by utilizing experts with varying capacities in a pixel-wise manner while necessitating substantially fewer FLOPs.
\end{itemize}

\section{Related Work}
\label{relatedworks}

\textbf{Implicit Neural Representation.} 
Significant advancements have been made in computer vision through the utilization of INR. 
These methods involve encoding objects into INRs and reconstructing them with a decoder, typically an MLP.
INRs have been applied to diverse tasks such as 3D modeling~\cite{Authors2, Authors3, Authors4, Authors5, Authors6, Authors7, Authors8, Authors9}, image generation~\cite{Authors10, Authors11}, SR~\cite{Authors13, Authors14, authors36, authors38}, and space-time video super-resolution~\cite{Authors28}.
Nevertheless, many works suffer from high computational complexity due to the enormous number of queries to the decoder.
Our study is dedicated to reducing the computational complexity of INR-based arbitrary-scale SR while preserving its efficacy through the appropriate allocation of capacity-varying decoders to each individual pixel. 
Although our approach may appear akin to KiloNeRF~\cite{Authors9}, it is imperative to discern a crucial distinction. KiloNeRF~\cite{Authors9} utilizes thousands of distinct MLPs for various segments of the scene, whereas our method effectively employs at most 4 depth-varying MLPs, reconstructing individual pixels regarding their specific reconstruction difficulties.

\textbf{Arbitrary-Scale Super-Resolution.}
Many SR researches~\cite{Authors1, Authors12, Authors15, Authors16, Authors17, authors34, authors35} were conducted to obtain rich representation power using sophisticated architectural designs and showed promising results.
However, these methods face limitations confined to fixed scale upscaling, necessitating multiple scale-specific models.
To address the issue, SR approaches with arbitrary upscale factor~\cite{Authors13, Authors14, authors31, authors32, authors33, authors36, authors37, authors38} have been introduced. 
MetaSR~\cite{authors32} pioneered upscaling images at arbitrary scales through a single trained models, followed by studies such as ArbSR~\cite{authors31} and SRWarp~\cite{authors33}. 
LIIF~\cite{Authors14} utilized implicit neural functions to upscale images to arbitrary scales. While this method demonstrated promising results, it encountered challenges in accurately preserving fine details when subjected to a large upscale factor. 
Subsequently, LTE~\cite{Authors13} was introduced with the aim of capturing intricate texture details within an image but also required a huge amount of computation since every output pixel needed to be queried using a single resource-intensive MLP decoder. 
More recent researches such as CLIT~\cite{authors36} and CiaoSR~\cite{authors38}, has improved the reconstruction quality of arbitrary-scale SR but requires excessive computational resources.
In contrast, our research introduces a novel approach that initially generates a map to specify which pixel should be processed by which one of the capacity-varying decoders. 
This meticulous assignment ensures that the INR is channeled through its most suitable decoder, thus allowing for highly accurate pixel value predictions while significantly reducing computational demands.

\textbf{Semantic-Aware Image Restoration.} 
Reconstructing texture details is crucial for better image restoration.
Studies such as \cite{authors39, authors40, authors41, authors42} have partitioned image regions and processed them with varying parameters, signifying a grwoing trend towards region-specific treatment.
More recently, ClassSR~\cite{Authors25} proposed difficulty-aware SR by using conventional SR networks with varying capacity to reduce the FLOPs required to reconstruct an image by attaching a class module. 
Although ClassSR~\cite{Authors25} demonstrated significant potential, it necessitated a three-step training procedure and relied on a pre-trained SR network's ability to accurately classify image regions based on their level of reconstruction difficulty. 
In contrast, our method uses a single-step training process for the entire network and does not rely on pre-trained SR networks. 
Additionally, while ClassSR~\cite{Authors25} classifies image patches, our approach classifies every pixel, leveraging the full range of complexity variations within image regions to handle diverse scenarios effectively.

\textbf{Mixture-of-Experts.} As the demand for computational efficiency increases, MoE has emerged as a promising strategy.
Its efficiency has led to its adoption across various domains, including natural language processing~\cite{authors43, authors45}, computer vision~\cite{authors44}, and multimodal applications~\cite{authors46}.
Despite its potential, MoE has not been widely adopted in SR tasks. Approaches resembling MoE, such as \cite{Authors25, authors30}, enhanced the performance of SR networks by employing multiple experts specialized in specific difficulty levels of patches. However, these approaches are limited to operating exclusively on a fixed scale factor. Furthermore, there is a noticeable absence of research dedicated to fully exploiting the potential of MoE in arbitrary-scale SR. Our work pioneers the utilization of MoE to tackle the computational complexity challenges inherent in arbitrary-scale SR, arising from its distinctive architectural characteristics.

\begin{figure*}
    \centering
    \includegraphics[width=\textwidth]{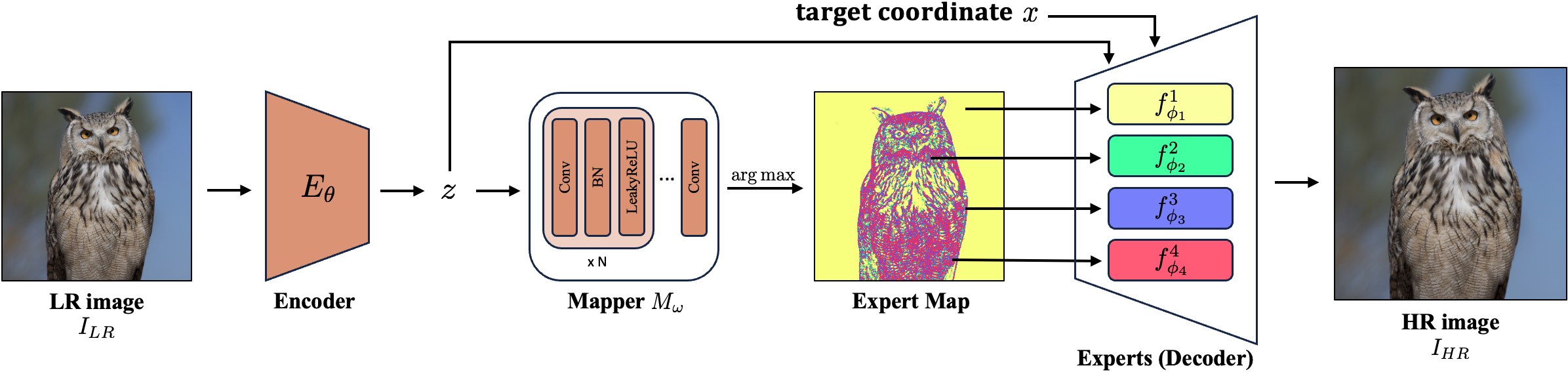}
    \vspace{-6mm}
    \caption{\textbf{Arbitrary-scale SR with MoEISR.} The INR (\(z\)) from the encoder (\(E_{\theta}\)) goes through the mapper (\(M_{\omega}\)), creating a expert map assigning the most suitable expert to each output pixel. Then, \(z\) and a target coordinate \(x\) are passed to the designated decoder (\(f^j_{\phi_j}\)) predicting the pixel's RGB value.}
    \vspace{-3mm}
    \label{fig:2}
\end{figure*}

\section{Proposed Method}
\label{proposedmethod}

\subsection{INR-Based SR}
INR-based SR networks such as LIIF~\cite{Authors14} and LTE~\cite{Authors13} extract the input image’s INR $z$, using conventional SR networks such as EDSR~\cite{Authors15}, RDN~\cite{Authors16}, and SwinIR~\cite{Authors17} as their encoders. This process can formally be expressed as 
\begin{equation}
    z=E_\theta(I_{LR}),
\end{equation}
where a LR input image $I_{LR} \in \mathbb R^{H\times W \times 3}$ is processed by $E_\theta$, an encoder parameterized by $\theta$, creating the INR $z\in \mathbb R^{H\times W \times D}$ which contains the information necessary to upscale $I_{LR}$. For each input pixel $i$, there is a corresponding INR $z_i \in \mathbb R^{D}$. 

To predict the RGB value of the output pixel $q$, a matching input pixel $k$ has to be found. 
Let us denote the center coordinate of pixel $q$ and $k$ as $x_q$ and $x_k$, respectively. 
These coordinates are relative to the image size, therefore the domain of $x_q$ and $x_k$ are equal (\eg $0 \le x_q, x_k \le 1$).
Within the relative coordinate space, an input pixel $k$ is chosen as the matching pixel for $q$ which has the closest Euclidean distance between $x_q$ and $x_k$ among all input pixels.

Finally, a decoder takes the representation vector $z_k$ and the relative coordinate $x_q-x_k$ to predict the final RGB value $s_q$ of the output pixel $q$:
\begin{equation}
    s_q=f_\phi(z_k,x_q-x_k),
\end{equation}
where $f_\phi$ denotes the decoder parameterized by $\phi$ .

\subsection{MoEISR}
INR-based SR models utilizes a single decoder $f_\phi$ to super-resolve images at arbitrary scales through varying the query positions $x_q$. However, this approach incurs substantial computational demand, as the decoder $f_\phi$ must be queried individually for each output pixel $q$. Given the inherent challenge of reducing the number of queries in INR-based SR models, we investigated a novel method to reduce the complexity of the decoding function. Our overall network architecture is illustrated in Fig.~\ref{fig:2}.

MoEISR utilizes a set of experts with diverse capacities by considering the reconstruction difficulty of each input pixel. 
Unlike traditional methods that rely on a single, resource-intensive decoder $f_\phi$, our approach utilizes a collection of decoders $f^1_{\phi_1}, f^2_{\phi_2}, ..., f^J_{\phi_J}$ with varying depths.
Among $J$ decoders, a mapper $M_\omega$ selects the most suitable expert for each input pixel by generating an expert map $D \in \mathbb R^{H \times W \times J}$ using the INR $z$:
\begin{equation}
    D = M_\omega(z).
\end{equation}
Specifically, $M_\omega$ assigns each pixel to an appropriate decoder, ensuring that computationally intensive processes are reserved for the most complex areas, such as object boundaries with abrupt RGB value transitions and regions with intricate textures.
A significant aspect of our approach is the classification performed in the latent space rather than the input space. We hypothesized that this would allow for fast yet accurate pixel-wise classification, as the latent representation $z_k$ contains sufficient information about its complexity and the requirements for upscaling. This hypothesis is supported by our experiments, which demonstrate that the expert map $D$ strongly correlates with local complexity (Fig.~\ref{fig:1}).
This correlation has profound implications for the performance of our model. By effectively matching pixel complexity with the appropriate decoder capacity, MoEISR not only improves the quality of image reconstruction but also optimizes computational efficiency. This strategic allocation of resources means that complex regions are handled with the necessary computational power, while simpler areas are processed more efficiently. Consequently, our approach significantly reduces the overall computational burden without compromising the quality of the reconstructed images.
 

Using the classification result from the mapper, each output pixel $q$, corresponding to input pixel $k$, is processed by the most suitable expert. This expert is selected by identifying the maximum value within the score vector $D_k \in \mathbb R^J$ of input pixel $k$, which represents the probability distribution for expert selection for the pixel's reconstruction. Mathematically, this process can be described as follows:
\begin{equation}
\label{eq:4}
    \begin{split}
    s_q = f^j_{\phi_j}(z_k, x_q-x_k),\\
    \text{where } j = \argmax_{i \in \{1, \dots, J\}} D_{ki},
    \end{split}
\end{equation}
and where $D_{ki}$ denotes $i$th element of $D_k$. 
Although the $\argmax$ operation is essential for achieving speed-up during test-time by selecting the most suitable expert, it is inherently non-differentiable, thereby obstructing gradient flow during training. To address this issue, we modify Eq.~\ref{eq:4} to employ a weighted sum of the decoders during training. This approach is formulated as follows:
\begin{equation}
    s_q = \sum_{j=1}^J g(D_k)_j \times f^j_{\phi_j}(z_k, x_q-x_k),
\end{equation}
where $g$ denotes the Gumbel-softmax function~\cite{Authors26, authors49}. 
Specifically, we employ the Gumbel-Softmax~\cite{Authors26, authors49} instead of the standard softmax to mitigate the bias of the mapper towards its early-stage decisions. The Gumbel-Softmax~\cite{Authors26, authors49} introduces controlled noise, thereby preventing overfitting and enhancing training robustness, as demonstrated in prior work~\cite{authors47, authors48}. This technique increases training flexibility and reduces the likelihood of the network converging to local optima. Importantly, we retain the original form of Eq.~\ref{eq:4} during inference to ensure optimal performance.

Our MoEISR framework is model-agnostic and compatible with any SR networks based on INR.
Furthermore, it is feasible to incorporate various techniques such as cell decoding and feature unfolding from LIIF~\cite{Authors14}, as well as scale-dependent phase estimation from LTE~\cite{Authors13}.
This flexibility allows for the incorporation of various enhancements to tailor MoEISR to specific tasks and requirements.

\begin{table}[t]
    \caption{Quantitative comparison on DIV2K validation set~\cite{Authors18}. \textbf{Bold} indicates the best PSNR among the backbone network and its MoEISR variant.}
    \label{tab:1}
    \centering\setlength\tabcolsep{2pt}%
    \resizebox{\textwidth}{!}{%
    \tiny{
    \begin{tabular}{c|ccc|ccccc||c|ccc|ccccc}
        \multirow{2}{*}{Method} & \multicolumn{3}{c|}{In-scale} & \multicolumn{5}{c||}{Out-of-scale} & \multirow{2}{*}{Method} & \multicolumn{3}{c|}{In-scale} & \multicolumn{5}{c}{Out-of-scale}\\
        & \(\times\)2 & \(\times\)3 & \(\times\)4 & \(\times\)6 & \(\times\)12 & \(\times\)18 & \(\times\)24 &\(\times\)30 & & \(\times\)2 & \(\times\)3 & \(\times\)4 & \(\times\)6 & \(\times\)12 & \(\times\)18 & \(\times\)24 &\(\times\)30\\ \hhline{==================}
        Bicubic & 31.01 & 28.22 & 26.66 & 24.82 &  22.27 & 21.00 &  20.19 & 19.59 & Bicubic & 31.01 & 28.22 & 26.66 & 24.82 &  22.27 & 21.00 &  20.19 & 19.59 \\
        EDSR-LIIF~\cite{Authors14} & \textbf{34.67} & 30.96 & 29.00 & 26.75 & \textbf{23.71} & \textbf{22.17} & \textbf{21.18} & 20.48 & EDSR-LTE~\cite{Authors13} & \textbf{34.72} & \textbf{31.02} & 29.04 & \textbf{26.81} & \textbf{23.78} & \textbf{22.23} & \textbf{21.24} & \textbf{20.53} \\ 
        EDSR-MoEISR(LIIF) -s & 34.60 & 30.91 & 28.96 & 26.70 & 23.66 & 22.13 & 21.14 & 20.45 & EDSR-MoEISR(LTE) -s & 34.71 & 31.01 & 29.04 & 26.80 & 23.76 & \textbf{22.23} & 21.23 & \textbf{20.53} \\
        EDSR-MoEISR(LIIF) -b & 34.66 & \textbf{30.98} & \textbf{29.01} & \textbf{26.76} & \textbf{23.71} & \textbf{22.17} & \textbf{21.18} & \textbf{20.49} & EDSR-MoEISR(LTE) -b & \textbf{34.72} & 31.01 & \textbf{29.05} & 26.80 & 23.77 & \textbf{22.23} & \textbf{21.24} & \textbf{20.53} \\  \hline
        RDN-LIIF~\cite{Authors14} & \textbf{34.99} & 31.26 & 29.27 & 26.99 & 23.89 & 22.34 & \textbf{21.31} & \textbf{20.59} & RDN-LTE~\cite{Authors13} & 35.04 & 31.32 & \textbf{29.33} & 27.04 & 23.95 & \textbf{22.40} & 21.36 & \textbf{20.64} \\
        RDN-MoEISR(LIIF) -s & 34.95 & 31.24 & 29.26 & 26.96 & 23.85 & 22.30 & 21.27 & 20.55 & RDN-MoEISR(LTE) -s & 35.03 & 31.32 & \textbf{29.33} & 27.04 & 23.95 & \textbf{22.40} & \textbf{21.37} & \textbf{20.64} \\        
        RDN-MoEISR(LIIF) -b  & \textbf{34.99} & \textbf{31.28} & \textbf{29.29} & \textbf{27.00} & \textbf{23.90} & \textbf{22.35} & \textbf{21.31} & \textbf{20.59} & RDN-MoEISR(LTE) -b & \textbf{35.05} & \textbf{31.33} & \textbf{29.33} & \textbf{27.05} & \textbf{23.96} & \textbf{22.40} & \textbf{21.37} & \textbf{20.64} \\ \hline
        SwinIR-LIIF~\cite{Authors13} & 35.17 & 31.46 & 29.46 & 27.15 & 24.02 & 22.43 & 21.40 & 20.67 & SwinIR-LTE~\cite{Authors13} & 35.24 & 31.50 & 29.51 & \textbf{27.20} & \textbf{24.09} & \textbf{22.50} & \textbf{21.47} & \textbf{20.73}\\ 
        SwinIR-MoEISR(LIIF) -s & 35.20 & 31.48 & 29.48 & 27.17 & 24.05 & \textbf{22.48} & 21.45 & 20.71 & SwinIR-MoEISR(LTE) -s & \textbf{35.25} & \textbf{31.51} & \textbf{29.52} & \textbf{27.20} & 24.08 & \textbf{22.50} & \textbf{21.47} & \textbf{20.73} \\
        SwinIR-MoEISR(LIIF) -b & \textbf{35.22} & \textbf{31.49} & \textbf{29.49} & \textbf{27.18} & \textbf{24.07} & \textbf{22.48} & \textbf{21.46} & \textbf{20.72} & SwinIR-MoEISR(LTE) -b & 35.24 & 31.50 & 29.51 & \textbf{27.20} & 24.08 & \textbf{22.50} & \textbf{21.47} & 20.72 \\
    \end{tabular}
    }}
    \vspace{-3mm}
\end{table}

\subsection{Loss Functions}
Our loss function has two distinct terms, $L_1$, the reconstruction loss, and $L_b$, the balance loss. 
$L_1$ serves to gauge the quality of reconstruction, which is commonly used in SR tasks. 
$L_b$ loss, on the other hand, helps the mapper to assign pixels to capacity-varying experts in a balanced manner. 
This dual loss contributes to the stable training of the network. 
The overall loss function can be described as:
\begin{equation}
  L = \alpha L_1 + \beta L_b,
  \label{eq:5}
\end{equation}
where $\alpha$ and $\beta$ represent the weights balancing two loss terms. 
As the hyperparameter $\alpha$ increases, the network converges towards prioritizing reconstruction quality. Conversely, as $\beta$ increases, the network tends to allocate experts more evenly, emphasizing computational efficiency over reconstruction quality.

The balance loss $L_b$ plays a critical role in MoEISR. 
Without $L_b$, the mapper might assign all input pixels to the deepest decoder, rather than utilizing all available experts in a balanced manner. 
The balance loss helps ensure a more balanced distribution of experts among pixels, promoting a more efficient utilization of the model's capacity.
One important point that we must put an emphasize on is that the overall loss function does not force the router to utilize all experts evenly.
Instead, it aids the router in assigning a more diverse range of experts, without necessitating an even distribution among them.
The formulation of the balance loss can be described as:
\begin{equation}
  L_b = \sum_{j=1}^J \bigg | w_j \sum_{k=1}^K D_{kj} - \frac{K}{J} \bigg|,
  \label{eq:6}
\end{equation}
where $J$ denotes the number of experts and $K$ represents the number of input pixels.
The balance loss is inspired by sparsely-gated mixture-of-experts~\cite{authors51} and ClassSR~\cite{Authors25}.
With $w_j=1$ for all $j$, it penalizes the mapper if its expert assignment is not uniform, ensuring $\mathbb E[D_k] \simeq \frac{1}{J}(1 \dots 1)^T$ for a random input pixel $k$. One can regulate the ratio of experts chosen by changing the value of $w_j$. This would be especially useful when the model is deployed to devices with varying computational capacities.


\begin{table*}[t]
    \caption{Quantitative comparison on various benchmark datasets. \textbf{Bold} indicates the best PSNR among the backbone network and its MoEISR variant.}
    \label{tab:2}
    \centering
    \setlength\tabcolsep{4pt}%
    \resizebox{\textwidth}{!}{%
    \begin{tabular}{c|ccccc|ccccc|ccccc|ccccc}
        \multirow{3}{*}{Method} & \multicolumn{5}{c|}{Set5~\cite{Authors19}} & \multicolumn{5}{c|}{Set14~\cite{Authors20}} & \multicolumn{5}{c|}{B100~\cite{Authors21}} & \multicolumn{5}{c}{Urban100~\cite{Authors22}} \\ \cline{2-21} 
        & \multicolumn{3}{c|}{In-scale} & \multicolumn{2}{c|}{Out-of-scale} & \multicolumn{3}{c|}{In-scale} & \multicolumn{2}{c|}{Out-of-scale} & \multicolumn{3}{c|}{In-scale} & \multicolumn{2}{c|}{Out-of-scale} & \multicolumn{3}{c|}{In-scale} & \multicolumn{2}{c}{Out-of-scale} \\
        & \(\times\)2 & \(\times\)3 & \multicolumn{1}{c|}{\(\times\)4} & \(\times\)6 & \(\times\)8 & \(\times\)2 & \(\times\)3 & \multicolumn{1}{c|}{\(\times\)4} & \(\times\)6 & \(\times\)8 & \(\times\)2 & \(\times\)3 & \multicolumn{1}{c|}{\(\times\)4} & \(\times\)6 & \(\times\)8 & \(\times\)2 & \(\times\)3 & \multicolumn{1}{c|}{\(\times\)4} & \(\times\)6 & \(\times\)8 \\ \hhline{=====================}    
        RDN~\cite{Authors16} & \textbf{38.24} & \textbf{34.71} & \multicolumn{1}{c|}{32.47} & - & - & \textbf{34.01} & \textbf{30.57} & \multicolumn{1}{c|}{28.81} & - & - & \textbf{32.34} & 29.26 & \multicolumn{1}{c|}{27.72} & - & - & 32.89 & 28.80 & \multicolumn{1}{c|}{26.61} & - & - \\
        RDN-LIIF~\cite{Authors14} & 38.17 & 34.68 & \multicolumn{1}{c|}{\textbf{32.50}} & 29.15 & 27.14 & 33.97 & 30.53 & \multicolumn{1}{c|}{28.80} & 26.64 & 25.15 & 32.32 & 29.26 & \multicolumn{1}{c|}{27.74} & \textbf{25.98} & 24.91 & 32.87 & 28.82 & \multicolumn{1}{c|}{26.68} & 24.20 & 22.79 \\
        RDN-MoEISR(LIIF) -s & 38.17 & 34.66 & \multicolumn{1}{c|}{32.49} & 29.20 & 27.13 &  33.98 & 30.52 & \multicolumn{1}{c|}{\textbf{28.82}} & 26.61 & 25.09 & 32.32 & 29.25 & \multicolumn{1}{c|}{27.74} & 25.97 & 24.91 & 32.82 & 28.80 & \multicolumn{1}{c|}{26.65} & 24.15 & 22.76 \\
        RDN-MoEISR(LIIF) -b & 38.19 & 34.69 & \multicolumn{1}{c|}{32.46} & \textbf{29.27} & \textbf{27.25} & 33.99 & 30.56 & \multicolumn{1}{c|}{\textbf{28.82}} & \textbf{26.66} & \textbf{25.20} & 32.33 & \textbf{29.28} & \multicolumn{1}{c|}{\textbf{27.75}} & \textbf{25.98} & \textbf{24.93} & \textbf{32.93} & \textbf{28.90} & \multicolumn{1}{c|}{\textbf{26.74}} & \textbf{24.24} & \textbf{22.84} \\ \hline
        RDN~\cite{Authors16} & 38.24 & 34.71 & \multicolumn{1}{c|}{32.47} & - & - & 34.01 & 30.57 & \multicolumn{1}{c|}{28.81} & - & - & 32.34 & 29.26 & \multicolumn{1}{c|}{27.72} & - & - & 32.89 & 28.80 & \multicolumn{1}{c|}{26.61} & - & - \\
        RDN-LTE~\cite{Authors13} & 38.23 & 34.72 & \multicolumn{1}{c|}{\textbf{32.61}} & \textbf{29.32} & \textbf{27.26} & 34.09 & 30.58 & \multicolumn{1}{c|}{\textbf{28.88}} & \textbf{26.71} & 25.16 & 32.36 & \textbf{29.30} & \multicolumn{1}{c|}{27.77} & \textbf{26.01} & \textbf{24.95} & 33.04 & 28.97 & \multicolumn{1}{c|}{26.81} & 24.28 & 22.88 \\
        RDN-MoEISR(LTE) -s & 38.21 & 34.70 & \multicolumn{1}{c|}{32.49} & 29.24 & 27.24 & 33.99 & \textbf{30.59} & \multicolumn{1}{c|}{28.86} & 26.69 & 25.18 & 32.36 & \textbf{29.30} & \multicolumn{1}{c|}{27.77} & 26.00 & 24.94 & 33.02 & 28.94 & \multicolumn{1}{c|}{26.79} & 24.25 & 22.86 \\
        RDN-MoEISR(LTE) -b & \textbf{38.25} & \textbf{34.78} & \multicolumn{1}{c|}{32.53} & 29.24 & 27.20 & \textbf{34.10} & 30.58 & \multicolumn{1}{c|}{28.86} & \textbf{26.71} & \textbf{25.22} & \textbf{32.37} & \textbf{29.30} & \multicolumn{1}{c|}{\textbf{27.78}} & \textbf{26.01} & \textbf{24.95} & \textbf{33.06} & \textbf{28.98} & \multicolumn{1}{c|}{\textbf{26.83}} & \textbf{24.31} & \textbf{22.89} \\ \hline
        SwinIR~\cite{Authors17} & \textbf{38.35} & \textbf{34.89} & \multicolumn{1}{c|}{32.72} & - & - & 34.14 & 30.77 & \multicolumn{1}{c|}{28.94} & - & - & \textbf{32.44} & \textbf{29.37} & \multicolumn{1}{c|}{27.83} & - & - & 33.40 & 29.29 & \multicolumn{1}{c|}{27.07} & - & - \\
        SwinIR-LIIF~\cite{Authors13} & 38.28 & 34.87 & \multicolumn{1}{c|}{32.73} & 29.46 & 27.36 & 34.14 & 30.75 & \multicolumn{1}{c|}{28.98} & 26.82 & 25.34 & 32.39 & 29.34 & \multicolumn{1}{c|}{27.84} & 26.07 & 25.01 & 33.36 & 29.33 & \multicolumn{1}{c|}{27.15} & 24.59 & 23.14 \\
        SwinIR-MoEISR(LIIF) -s & 38.30 & 34.85 & \multicolumn{1}{c|}{32.72} & 29.45 & 27.34 & \textbf{34.25} & 30.77 & \multicolumn{1}{c|}{29.01} & 26.82 & 25.35 & 32.42 & 29.36 & \multicolumn{1}{c|}{27.84} & 26.07 & 25.01 & 33.41 & 29.40 & \multicolumn{1}{c|}{27.19} & 24.58 & 23.13 \\
        SwinIR-MoEISR(LIIF) -b & 38.30 & 34.85 & \multicolumn{1}{c|}{\textbf{32.77}} & \textbf{29.48} & \textbf{27.39} & 34.24 & \textbf{30.78} & \multicolumn{1}{c|}{\textbf{29.02}} & \textbf{26.87} & \textbf{25.36} & 32.43 & \textbf{29.37} & \multicolumn{1}{c|}{\textbf{27.85}} & \textbf{26.08} & \textbf{25.02} & \textbf{33.48} & \textbf{29.42} & \multicolumn{1}{c|}{\textbf{27.22}} & \textbf{24.62} & \textbf{23.16} \\ \hline
        SwinIR-LTE~\cite{Authors13} & 38.33 & 34.89 & \multicolumn{1}{c|}{\textbf{32.81}} & 29.50 & 27.35 & 34.25 & \textbf{30.80} & \multicolumn{1}{c|}{\textbf{29.06}} & 26.86 & \textbf{25.42} & 32.44 & \textbf{29.39} & \multicolumn{1}{c|}{27.86} & \textbf{26.09} & \textbf{25.03} & 33.50 & 29.41 & \multicolumn{1}{c|}{27.24} & 24.62 & 23.17 \\
        SwinIR-MoEISR(LTE) -s & \textbf{38.35} & \textbf{34.90} & \multicolumn{1}{c|}{32.78} & \textbf{29.53} & \textbf{27.42} & \textbf{34.27} & \textbf{30.80} & \multicolumn{1}{c|}{29.03} & \textbf{26.87} & 25.40 & \textbf{32.45} & \textbf{29.39} & \multicolumn{1}{c|}{\textbf{27.87}} & \textbf{26.09} & \textbf{25.03} & \textbf{33.52} & \textbf{29.45} & \multicolumn{1}{c|}{\textbf{27.25}} & \textbf{24.66} & \textbf{23.18} \\
        SwinIR-MoEISR(LTE) -b & 38.34 & 34.88 & \multicolumn{1}{c|}{32.79} & 29.47 & 27.37 & 34.24 & 30.77 & \multicolumn{1}{c|}{29.03} & 26.81 & 25.39 & 32.44 & 29.38 & \multicolumn{1}{c|}{27.86} & \textbf{26.09} & \textbf{25.03} & 33.49 & 29.40 & \multicolumn{1}{c|}{27.24} & 24.63 & \textbf{23.18}
    \end{tabular}%
    }
    \vspace{-1mm}
\end{table*}
\begin{table*}[t]
    \caption{FLOPs comparison on Set14 validation set~\cite{Authors20}. Given that the encoder's FLOPs of all methods are identical, only the decoder's FLOPs are reported. The decoder in MoEISR consists of a router and a set of experts, while LIIF~\cite{Authors14} comprises a single MLP decoder. Set14~\cite{Authors20} images were utilized as LR images and upscaled with various scales. \textbf{Bold} indicates the least FLOPs required among the baseline network and its MoEISR variant.}
    \label{tab:3}
    \centering\setlength\tabcolsep{3pt}%
    \scriptsize{
    \resizebox{\textwidth}{!}{%
    \begin{tabular}{c|cc|cc|cc|cc|cc|cc|cc|cc}
        \multirow{2}{*}{Method} & \multicolumn{6}{c|}{In-scale} & \multicolumn{10}{c}{Out-of-scale}\\
        & \(\times\)2 & \% & \(\times\)3 & \% & \(\times\)4 & \% & \(\times\)6 & \% & \(\times\)12 & \% & \(\times\)18 & \% & \(\times\)24 & \% & \(\times\)30 & \% \\ \hhline{=================}
        EDSR-LIIF~\cite{Authors14} & 1.27TF & 100\% & 2.87TF & 100\% & 5.10TF & 100\% & 11.47TF & 100\% & 45.86TF & 100\% & 103.18TF & 100\% & 183.44TF & 100\% & 286.62TF & 100\%  \\ 
        EDSR-MoEISR(LIIF) -b & 0.87TF & 68.5\% & 1.95TF & 67.9\% & 3.47TF & 68.0\% & 7.80TF & 68.0\% & 31.17TF & 68.0\% & 70.12TF & 68.0\% & 124.68TF & 68.0\% & 194.68TF & 68.0\% \\
        EDSR-MoEISR(LIIF) -s & \textbf{0.34TF} & \textbf{26.7\%} & \textbf{0.76TF} & \textbf{26.5\%} & \textbf{1.34TF} & \textbf{26.3\%} & \textbf{3.01TF} & \textbf{26.2\%} & \textbf{12.02TF} & \textbf{26.2\%} & \textbf{27.03TF} & \textbf{26.2\%} & \textbf{48.06TF} & \textbf{26.2\%} & \textbf{75.09TF} & \textbf{26.2\%} \\ 
        \hline
        EDSR-LTE~\cite{Authors13} & 0.73TF & 100\% & 1.64TF & 100\% & 2.91TF & 100\% & 6.54TF & 100\% & 26.17TF & 100\% & 58.89TF & 100\% & 104.69TF & 100.0\% & 163.57TF & 100\%  \\
        EDSR-MoEISR(LTE) -b & 0.67TF & 91.8\% & 1.51TF & 92.1\% & 2.67TF & 91.8\% & 6.01TF & 91.9\% & 24.01TF & 91.7\% & 54.01TF & 91.7\% & 96.01TF & 91.7\% & 150.02TF & 91.7\% \\
        EDSR-MoEISR(LTE) -s & \textbf{0.20TF} & \textbf{27.4\%} & \textbf{0.44TF} & \textbf{26.8\%} & \textbf{0.79TF} & \textbf{27.1\%} & \textbf{1.76TF} & \textbf{26.9\%} & \textbf{7.02TF} & \textbf{26.8\%} & \textbf{15.80TF} & \textbf{26.8\%} & \textbf{28.08TF} & \textbf{26.8\%} & \textbf{43.88TF} & \textbf{26.8\%} \\ \hline
        RDN-LIIF~\cite{Authors14} & 1.27TF & 100\% & 2.86TF & 100\% & 5.09TF & 100\% & 11.46TF & 100\% & 45.86TF & 100\% & 103.18TF & 100\% & 183.43TF & 100\% & 286.62TF & 100\% \\         
        RDN-MoEISR(LIIF) -b  & 0.99TF & 78.0\% & 2.21TF & 77.3\% & 3.94TF & 77.4\% & 8.85TF & 77.2\% & 35.41TF & 77.2\% & 79.67TF & 77.2\% & 141.61TF & 77.2\% & 221.29TF & 77.2\% \\
        RDN-MoEISR(LIIF) -s & \textbf{0.34TF} & \textbf{27.0\%} & \textbf{0.77TF} & \textbf{27.0\%} & \textbf{1.37TF} & \textbf{27.0\%} & \textbf{3.08TF} & \textbf{26.9\%} & \textbf{12.32TF} & \textbf{26.9\%} & \textbf{27.71TF} & \textbf{26.9\%} & \textbf{49.26TF} & \textbf{26.9\%} & \textbf{76.96TF} & \textbf{26.9\%} \\ \hline
        RDN-LTE~\cite{Authors13} & 0.72TF & 100\% & 1.63TF & 100\% & 2.90TF & 100\% & 6.54TF & 100\% & 26.17TF & 100\% & 58.88TF & 100\% & 104.68TF & 100\% & 163.57TF & 100\% \\
        RDN-MoEISR(LTE) -b & 0.69TF & 95.9\% & 1.54TF & 94.5\% & 2.74TF & 94.5\% & 6.16TF & 94.2\% & 24.64TF & 94.2\% & 55.43TF & 94.1\% & 98.54TF & 94.1\% & 153.97TF & 94.1\% \\
        RDN-MoEISR(LTE) -s & \textbf{0.23TF} & \textbf{32.3\%} & \textbf{0.52TF} & \textbf{32.1\%} & \textbf{0.93TF} & \textbf{32.2\%} & \textbf{2.08TF} & \textbf{31.8\%} & \textbf{8.33TF} & \textbf{31.8\%} & \textbf{18.75TF} & \textbf{31.8\%} & \textbf{33.32TF} & \textbf{31.8\%} & \textbf{52.07TF} & \textbf{31.8\%} \\ \hline
        SwinIR-LIIF~\cite{Authors13} & 1.27TF & 100\% & 2.86TF & 100\% & 5.09TF & 100\% & 11.46TF & 100\% & 45.86TF & 100\% & 103.18TF & 100\% & 183.44TF & 100\% & 286.62TF & 100\% \\ 
        SwinIR-MoEISR(LIIF) -b & 0.95TF & 74.8\% & 2.12TF & 74.1\% & 3.77TF & 74.1\% & 8.47TF & 73.9\% & 33.89TF & 73.9\% & 76.22TF & 73.9\% & 135.51TF & 73.9\% & 211.49TF & 73.8\% \\ 
        SwinIR-MoEISR(LIIF) -s  & \textbf{0.35TF} & \textbf{27.7\%} & \textbf{0.78TF} & \textbf{27.3\%} & \textbf{1.39TF} & \textbf{27.3\%} & \textbf{3.13TF} & \textbf{27.3\%} & \textbf{12.49TF} & \textbf{27.2\%} & \textbf{28.10TF} & \textbf{27.2\%} & \textbf{49.99TF} & \textbf{27.3\%} & \textbf{78.00TF} & \textbf{27.2\%} \\ \hline
        SwinIR-LTE~\cite{Authors13} & 0.72TF & 100\% & 1.63TF & 100\% & 2.91TF & 100\% & 6.54TF & 100\% & 26.17TF & 100\% & 58.88TF & 100\% & 104.68TF & 100\% & 163.57TF & 100\% \\ 
        SwinIR-MoEISR(LTE) -b & 0.67TF & 93.1\% & 1.53TF & 93.9\% & 2.70TF & 92.8\% & 6.03TF & 92.2\% & 24.07TF & 92.0\% & 54.07TF & 91.8\% & 96.10TF & 91.8\% & 150.16TF & 91.8\% \\ 
        SwinIR-MoEISR(LTE) -s & \textbf{0.25TF} & \textbf{34.9\%} & \textbf{0.54TF} & \textbf{33.2\%} & \textbf{0.94TF} & \textbf{32.4\%} & \textbf{2.07TF} & \textbf{31.7\%} & \textbf{8.21TF} & \textbf{31.4\%} & \textbf{18.42TF} & \textbf{31.3\%} & \textbf{32.75TF} & \textbf{31.3\%} & \textbf{51.16TF} & \textbf{31.3\%} \\ 
    \end{tabular}%
    }
    }
\vspace{-2EX}
\end{table*}

\subsection{Training Strategy} 
For training, we first downscale the input image to a random scale from \(1\times\) to \(4\times\) so that the network can learn to upscale the image in an arbitrary scale. 
Similar to the evaluation methodologies employed in LIIF~\cite{Authors14} and LTE~\cite{Authors13}, our network is also evaluated at scales like \(8\times\) and \(32 \times\) that have not been encountered during its training phase with the aim of demonstrating the network's capacity for generalization.
During the training phase, we aggregate the outputs of individual experts, weighted by probabilities derived from Gumbel-softmax~\cite{Authors26, authors49}, indicating the suitability of each decoder for an output pixel. In the testing phase, we simplify the methodology by straightforwardly selecting the expert with the highest probability, leading to the computation of the final output. This streamlined process achieves image reconstruction with significantly reduced computational load.

\section{Experiment}

The code implementation of MoEISR will be available upon acceptance. More details and additional results can be found in our supplementary material.

\textbf{Datasets.} Our approach employs the identical dataset as our backbone models: LIIF~\cite{Authors14} and LTE~\cite{Authors13}. 
Both of these models are trained on the DIV2K dataset from the NTIRE 2017 Challenge~\cite{Authors18}. 
Subsequently, our evaluation procedure is based on several benchmark datasets, including the DIV2K validation set~\cite{Authors18}, Set5~\cite{Authors19}, Set14~\cite{Authors20}, B100~\cite{Authors21}, and Urban100~\cite{Authors22}.

\textbf{Implementation Details.} Since our research is model-agnostic and generally applicable to conventional INR-based arbitrary-scale SR networks, we adopt a consistent approach with the implementations of our backbone models. Specifically, we configure our network to process \(48 \times 48\) patches as input data. For the baseline encoders, we choose EDSR~\cite{Authors15}, RDN~\cite{Authors16}, and SwinIR~\cite{Authors17}.
For MoEISR employing LIIF~\cite{Authors14} as the backbone, we employ 4 capacity-varying experts, while for the LTE~\cite{Authors13} version, 3 capacity-varying experts are utilized, each mirroring the depth of its original decoder as the heaviest expert.
We use the conventional L1 loss~\cite{Authors25} and the balance loss with \(\alpha = 3000\) and \(\beta = 1\) with the Adam~\cite{Authors24} optimizer. The learning rate and the epochs are set equivalent to that of the backbone network.
For the mapper, we use a 5-layer convolutional neural network and a Gumbel-softmax~\cite{Authors26, authors49} with temperature hyperparameter \(\tau = 1\) for normalization by default.

\subsection{Evaluation}
\textbf{Quantitative Results.} Tab.~\ref{tab:1} describes a quantitative analysis between the MoEISR approach based on LIIF~\cite{Authors14} and LTE~\cite{Authors13}, using EDSR~\cite{Authors15}, RDN~\cite{Authors16}, and SwinIR~\cite{Authors17} as encoder. 
MoEISR -b is our baseline model, which incorporates experts with 256 hidden dimensions while varying the depth in the decoder and MoEISR -s employs experts with 128 hidden dimensions with varying depths, notably decreasing computational complexity.
Tab.~\ref{tab:1} clearly indicates that the MoEISR -b models exhibit impressive performance, attaining comparable or even superior PSNR values to its backbone network across various scenarios.
Interestingly, MoEISR -s also demonstrates competitive reconstruction capabilities, occasionally even outperforming the original backbone network. 
In Tab.~\ref{tab:2}, we further conduct a comparative analysis on MoEISR based on LIIF~\cite{Authors14} and LTE~\cite{Authors13}, using RDN~\cite{Authors16} and SwinIR~\cite{Authors17} as encoders, across various benchmark datasets. 
As already shown in Tab.~\ref{tab:1}, MoEISR -b outperforms its respective backbone networks in the majority of cases and MoEISR -s also demonstrates impressive performance, occasionally even surpassing MoEISR -b.
An intriguing observation that drew our attention was the PSNR performance of SwinIR-MoEISR(LTE) -s model, surpassing not only the SwinIR-LTE~\cite{Authors13} model but also the MoEISR(LTE) -b model. We attribute this outcome to the superior representational capabilities of SwinIR~\cite{Authors17}, which significantly outperform those of EDSR~\cite{Authors15} and RDN~\cite{Authors16}.

\begin{figure*}
\setlength\tabcolsep{2pt}%
\scriptsize{
\begin{tabular}{cccccccccc}
   \rotatebox{90}{\phantom{aaaa}\(\times\)3.3} &
   \includegraphics[ width=0.097\linewidth, height=\linewidth, keepaspectratio]{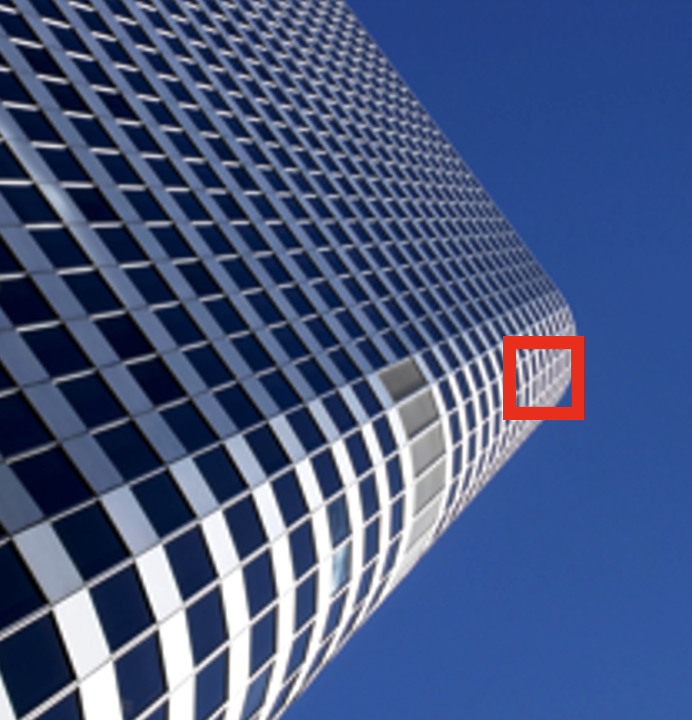} &
   \includegraphics[ width=0.097\linewidth, height=\linewidth, keepaspectratio]{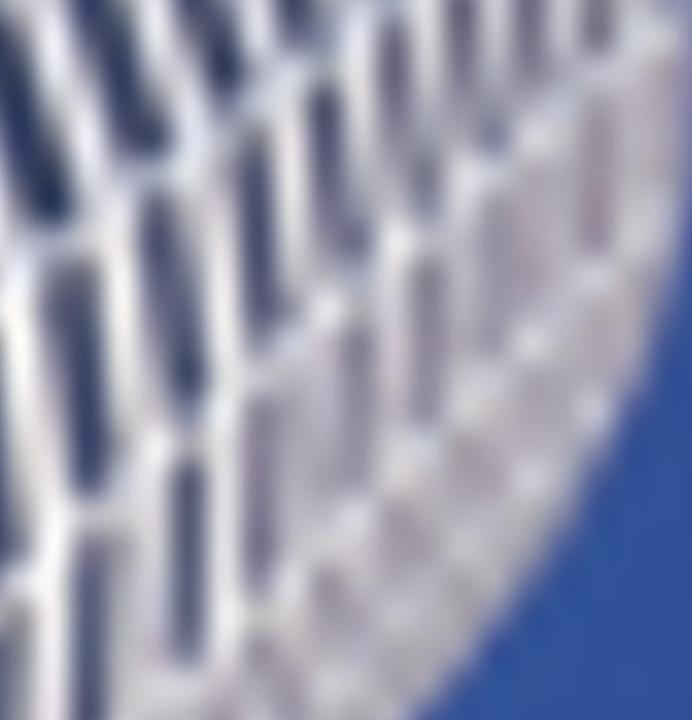} &
   \includegraphics[ width=0.097\linewidth, height=\linewidth, keepaspectratio]{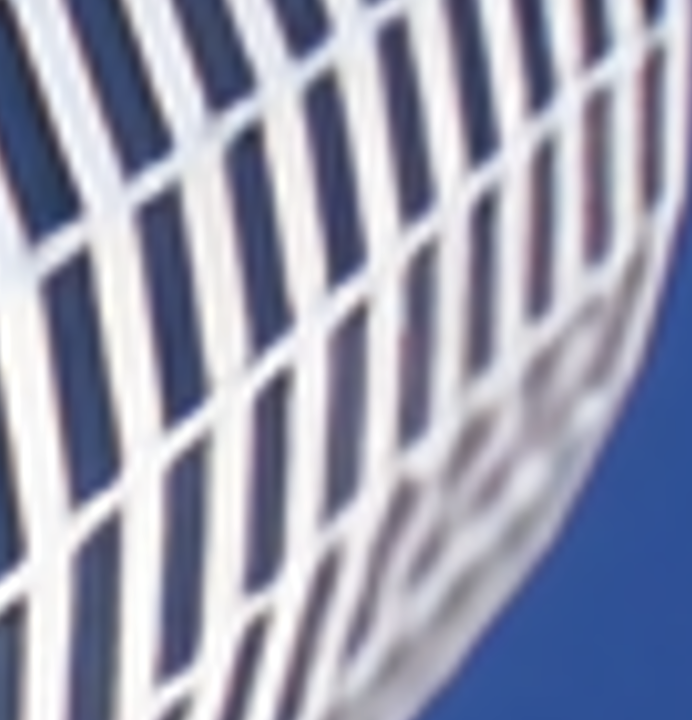} &
   \includegraphics[ width=0.097\linewidth, height=\linewidth, keepaspectratio]{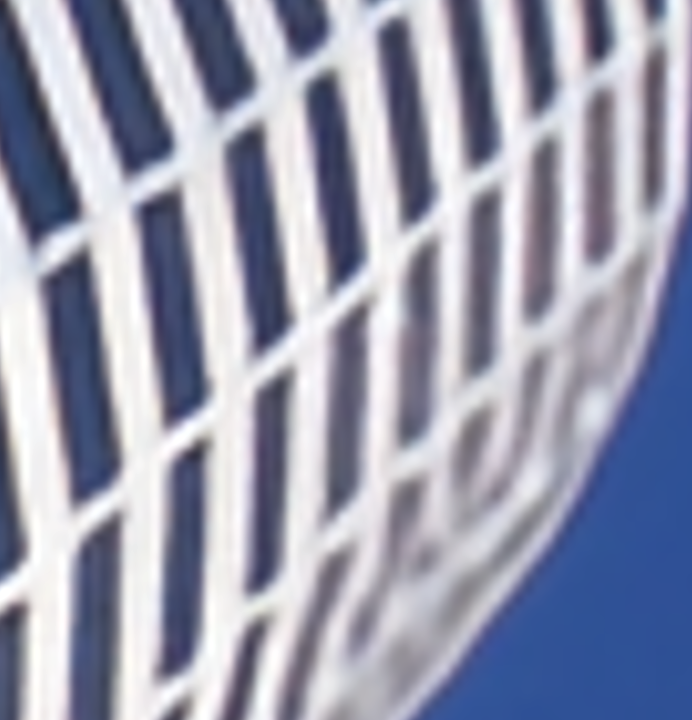} &
   \includegraphics[ width=0.097\linewidth, height=\linewidth, keepaspectratio]{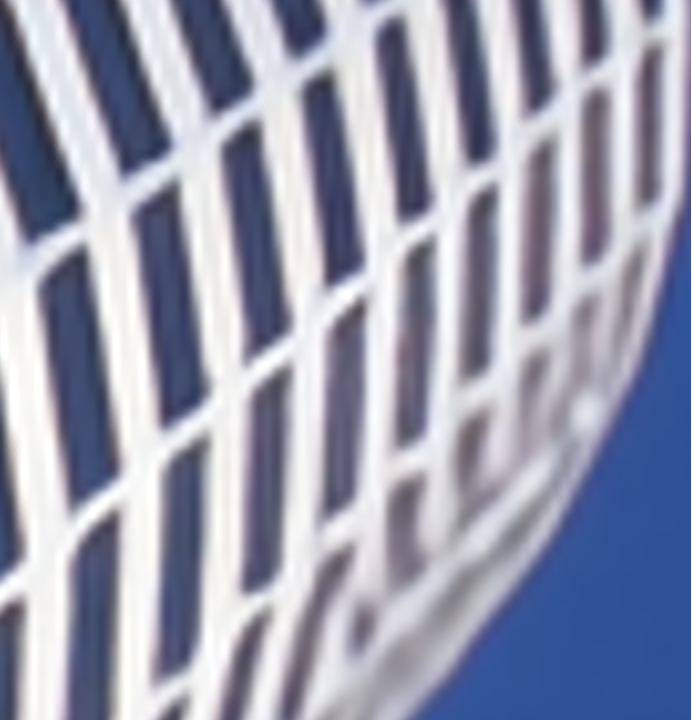} &
   \includegraphics[ width=0.097\linewidth, height=\linewidth, keepaspectratio]{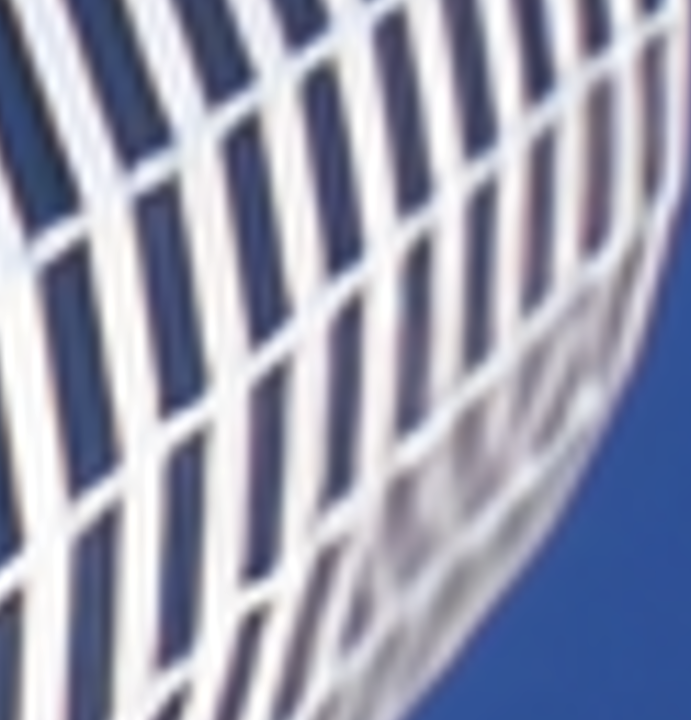} &
   \includegraphics[ width=0.097\linewidth, height=\linewidth, keepaspectratio]{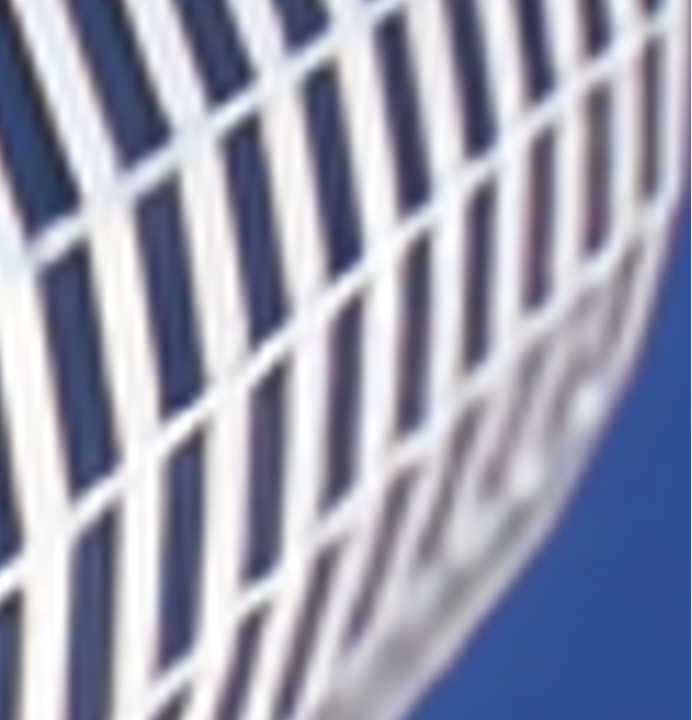} &
   \includegraphics[ width=0.097\linewidth, height=\linewidth, keepaspectratio]{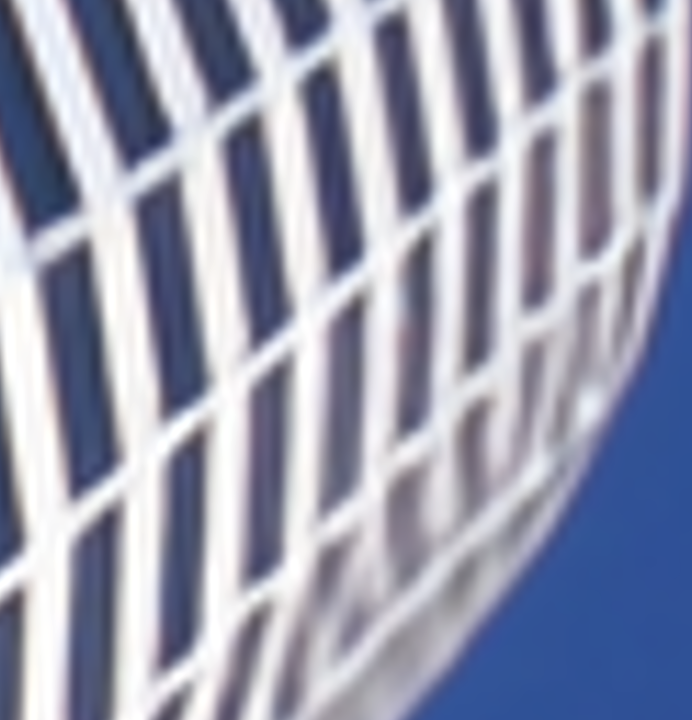} &
   \includegraphics[ width=0.097\linewidth, height=\linewidth, keepaspectratio]{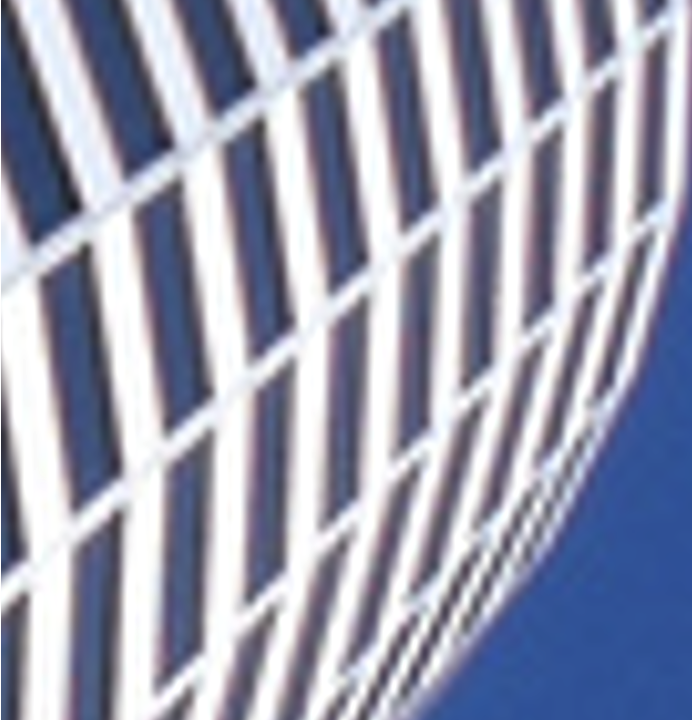}\\
\rotatebox{90}{\phantom{aaaa}\(\times\)4} &
   \includegraphics[ width=0.097\linewidth, height=\linewidth, keepaspectratio]{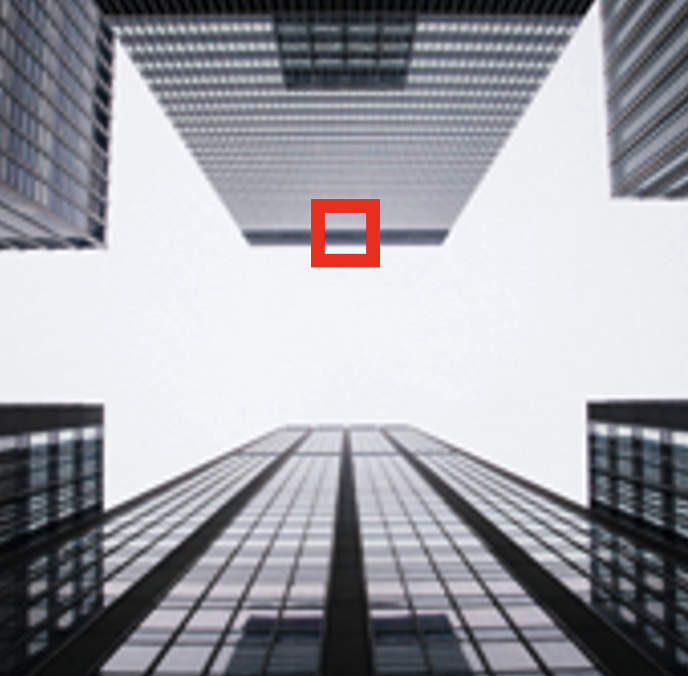} &
   \includegraphics[ width=0.097\linewidth, height=\linewidth, keepaspectratio]{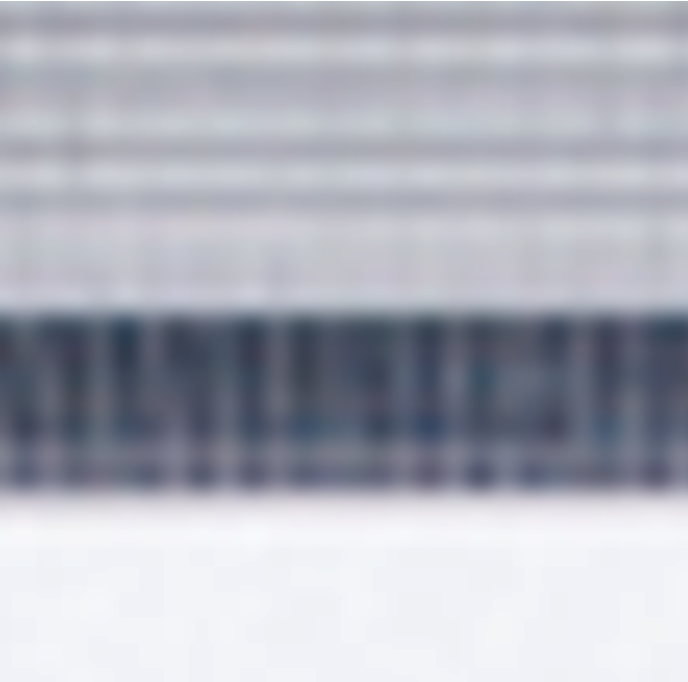} &
   \includegraphics[ width=0.097\linewidth, height=\linewidth, keepaspectratio]{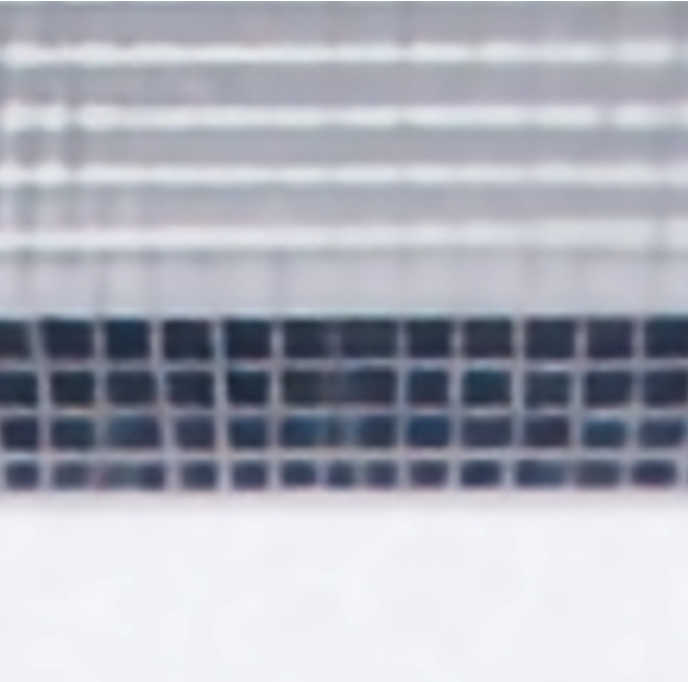} &
   \includegraphics[ width=0.097\linewidth, height=\linewidth, keepaspectratio]{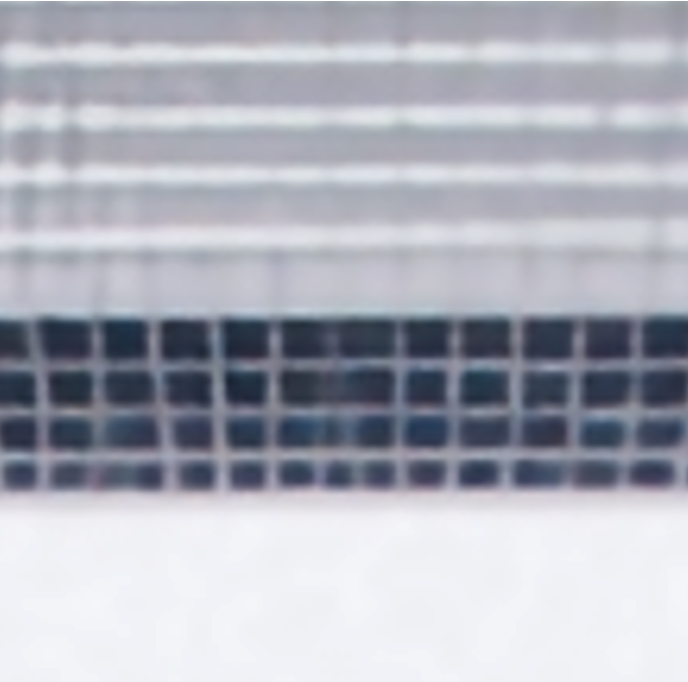} &
   \includegraphics[ width=0.097\linewidth, height=\linewidth, keepaspectratio]{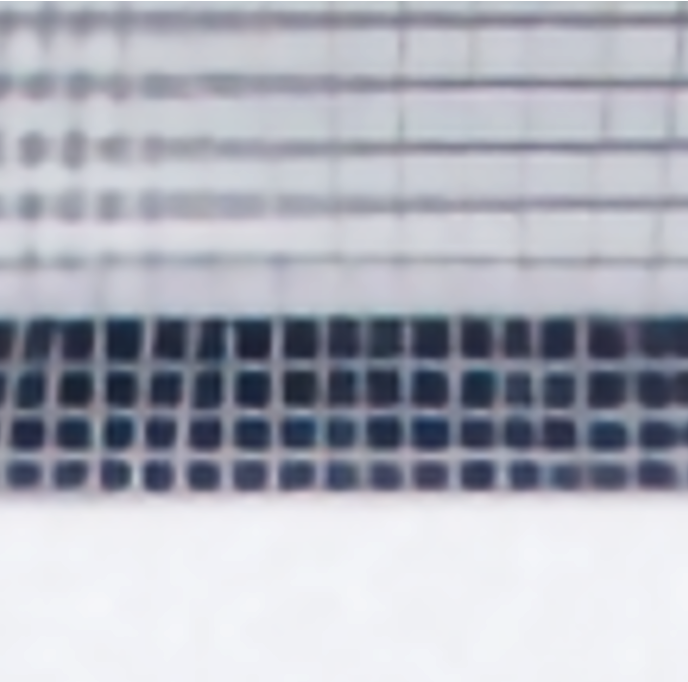} &
   \includegraphics[ width=0.097\linewidth, height=\linewidth, keepaspectratio]{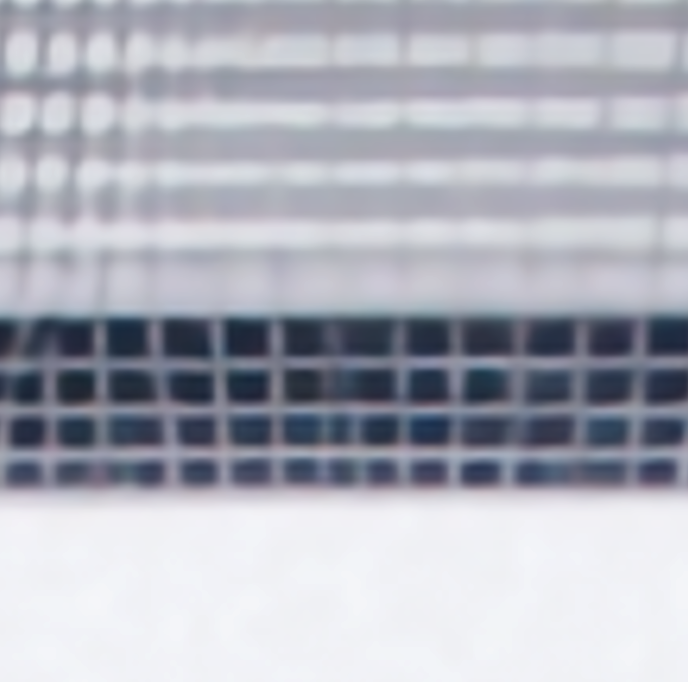} &
   \includegraphics[ width=0.097\linewidth, height=\linewidth, keepaspectratio]{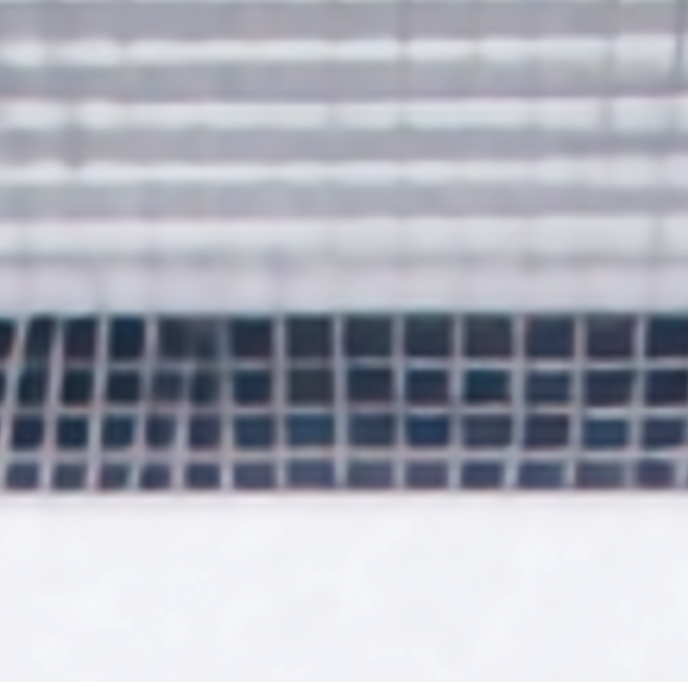} &
   \includegraphics[ width=0.097\linewidth, height=\linewidth, keepaspectratio]{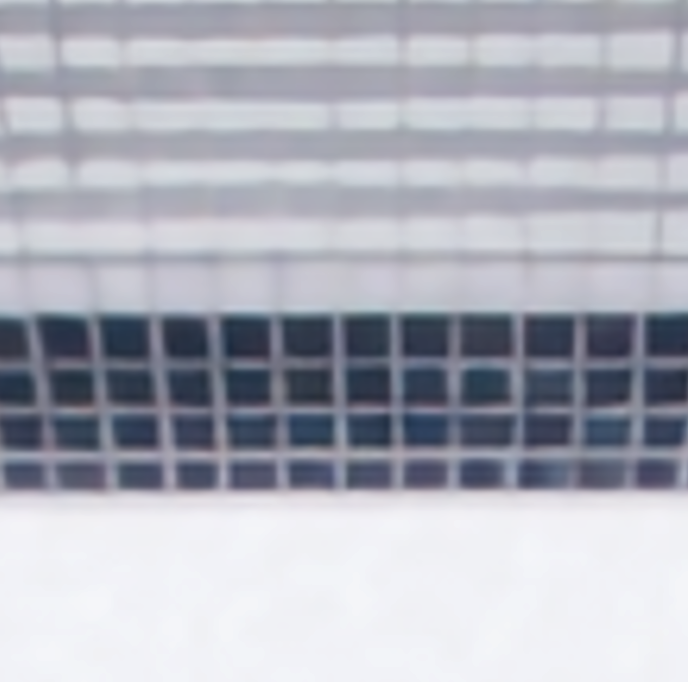} &
   \includegraphics[ width=0.097\linewidth, height=\linewidth, keepaspectratio]{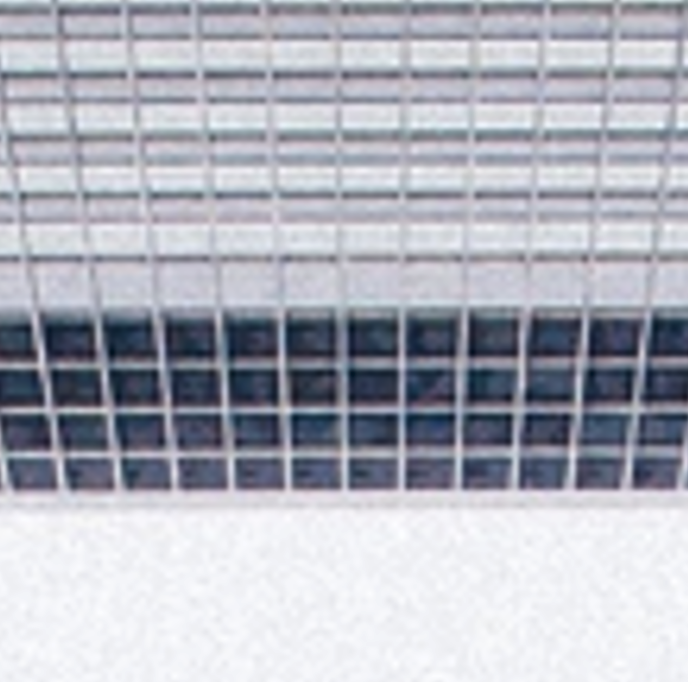}\\
\rotatebox{90}{\phantom{aaaa}\(\times\)8} &
   \includegraphics[ width=0.097\linewidth, height=\linewidth, keepaspectratio]{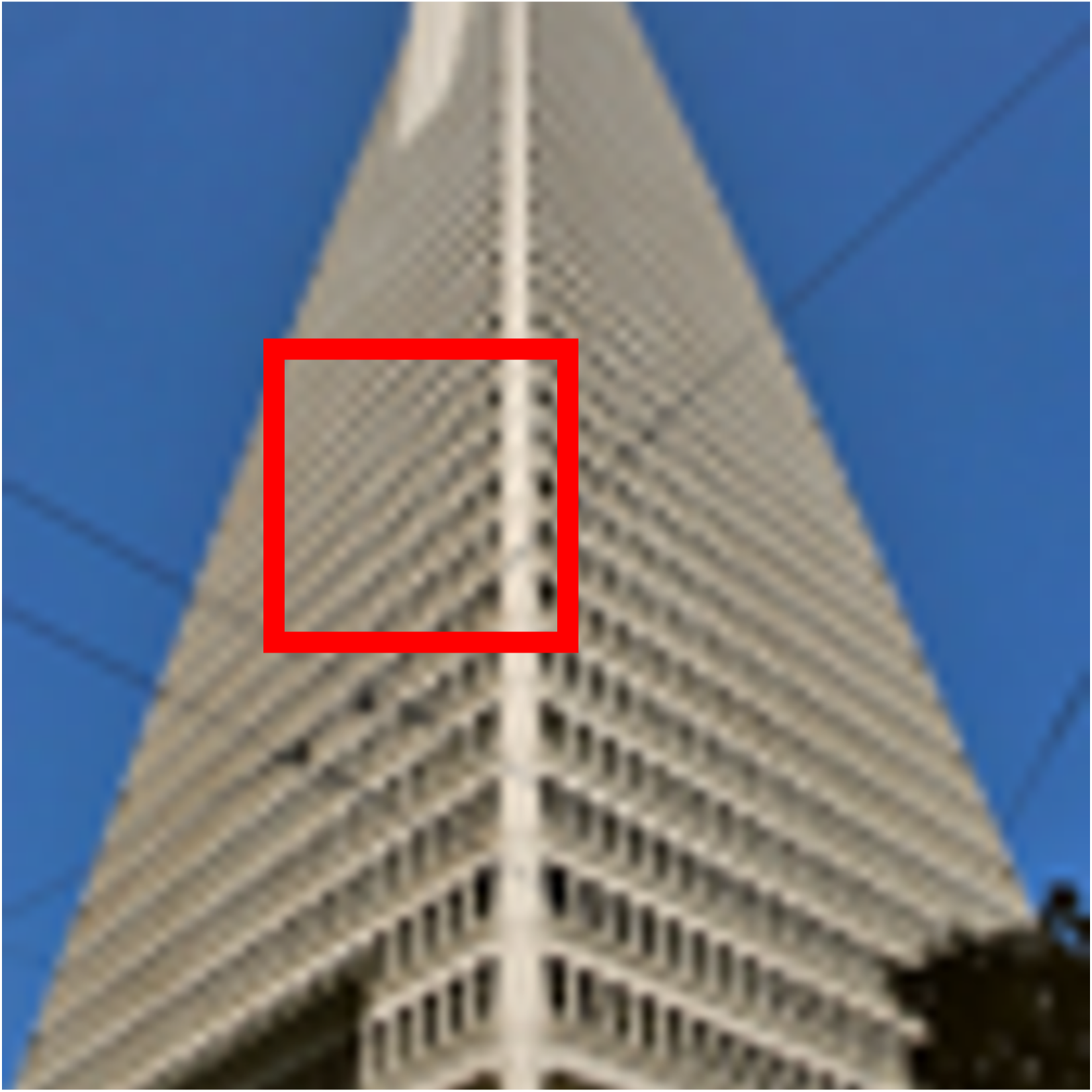} &
   \includegraphics[ width=0.097\linewidth, height=\linewidth, keepaspectratio]{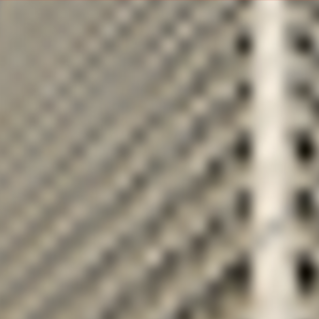} &
   \includegraphics[ width=0.097\linewidth, height=\linewidth, keepaspectratio]{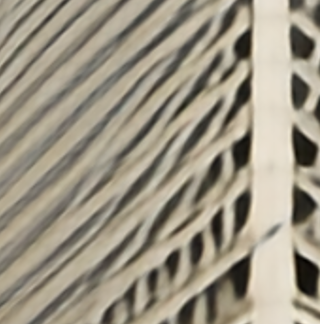} &
   \includegraphics[ width=0.097\linewidth, height=\linewidth, keepaspectratio]{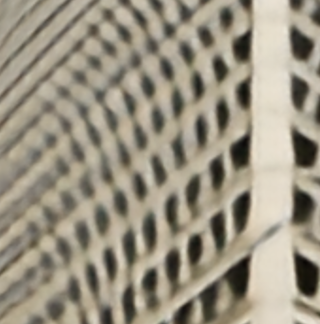} &
   \includegraphics[ width=0.097\linewidth, height=\linewidth, keepaspectratio]{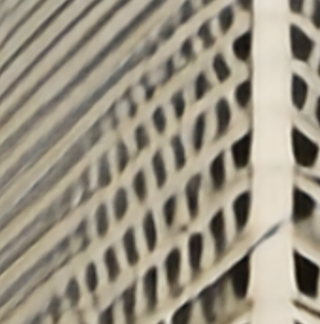} &
   \includegraphics[ width=0.097\linewidth, height=\linewidth, keepaspectratio]{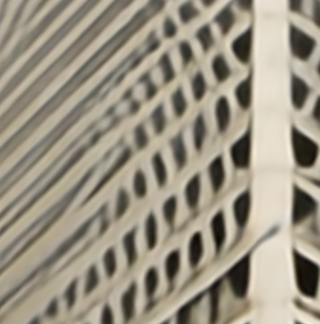} &
   \includegraphics[ width=0.097\linewidth, height=\linewidth, keepaspectratio]{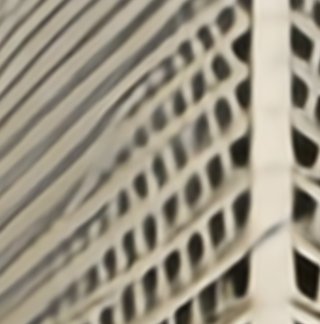} &
   \includegraphics[ width=0.097\linewidth, height=\linewidth, keepaspectratio]{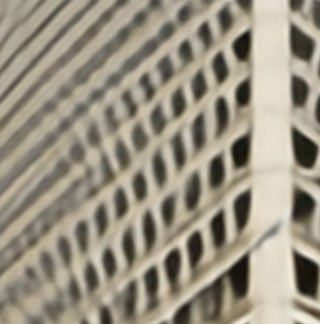} &
   \includegraphics[ width=0.097\linewidth, height=\linewidth, keepaspectratio]{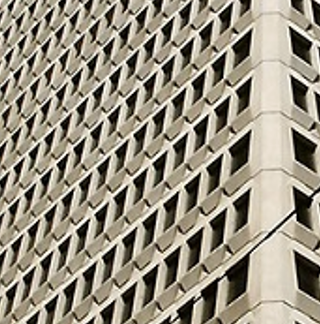}\\
\rotatebox{90}{\phantom{aaaa}\(\times\)18} &
   \includegraphics[ width=0.097\linewidth, height=\linewidth, keepaspectratio]{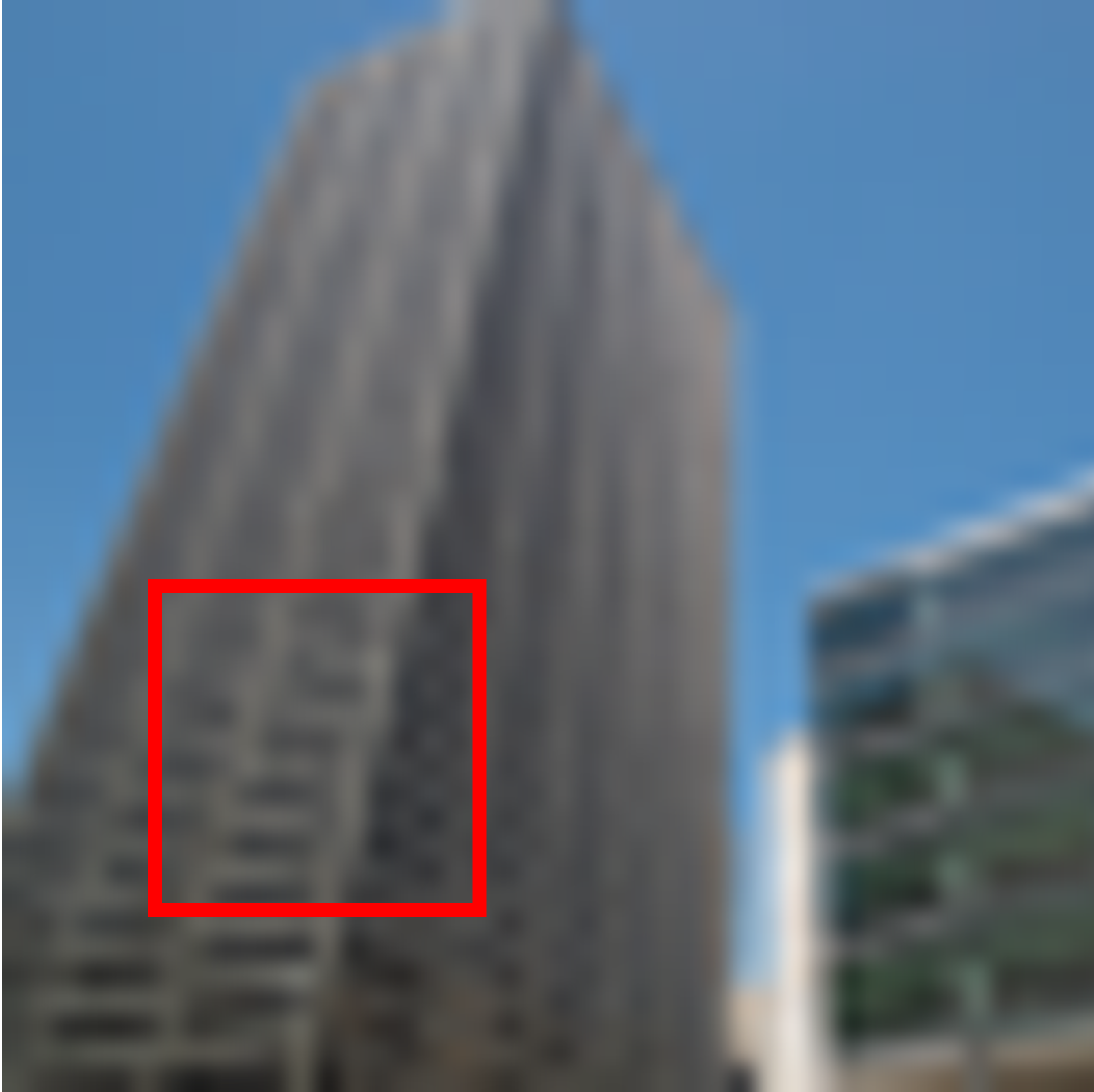} &
   \includegraphics[ width=0.097\linewidth, height=\linewidth, keepaspectratio]{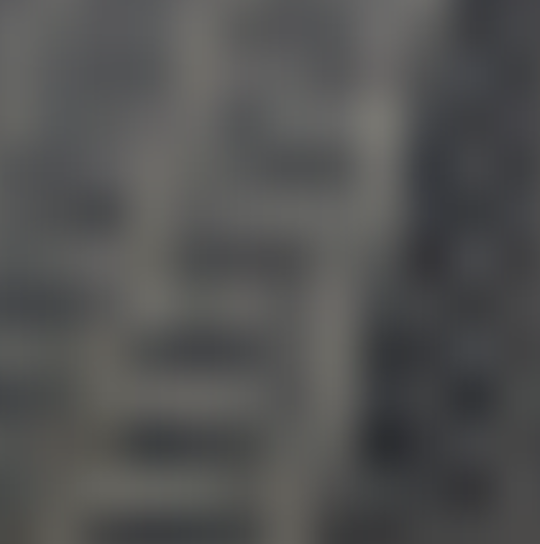} &
   \includegraphics[ width=0.097\linewidth, height=\linewidth, keepaspectratio]{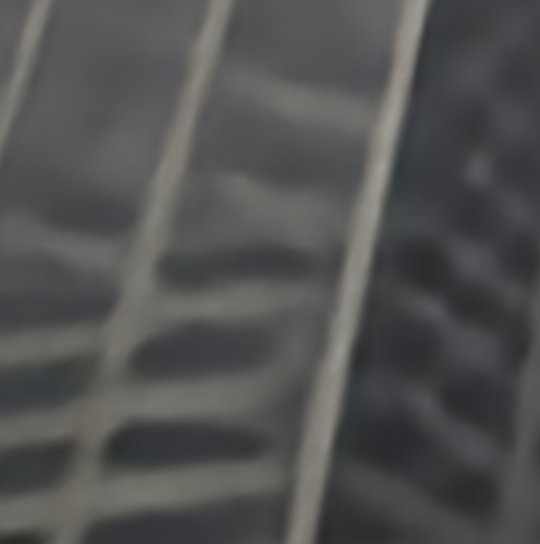} &
   \includegraphics[ width=0.097\linewidth, height=\linewidth, keepaspectratio]{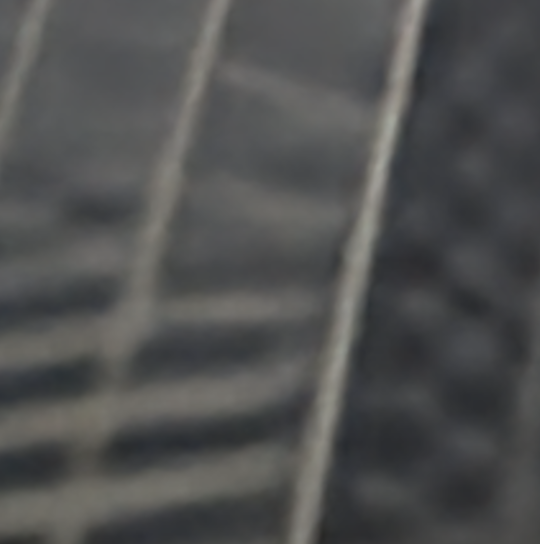} &
   \includegraphics[ width=0.097\linewidth, height=\linewidth, keepaspectratio]{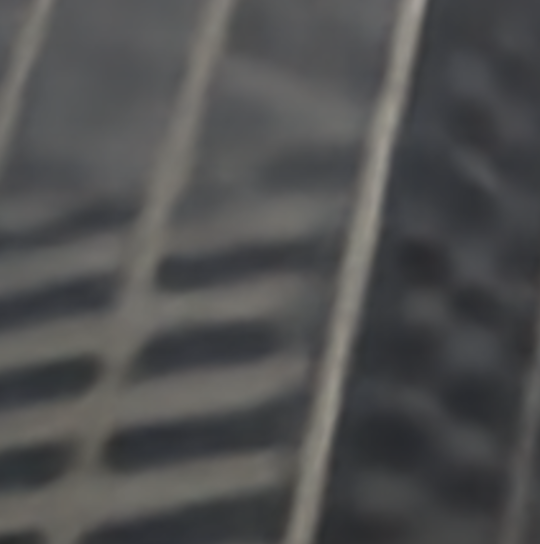} &
   \includegraphics[ width=0.097\linewidth, height=\linewidth, keepaspectratio]{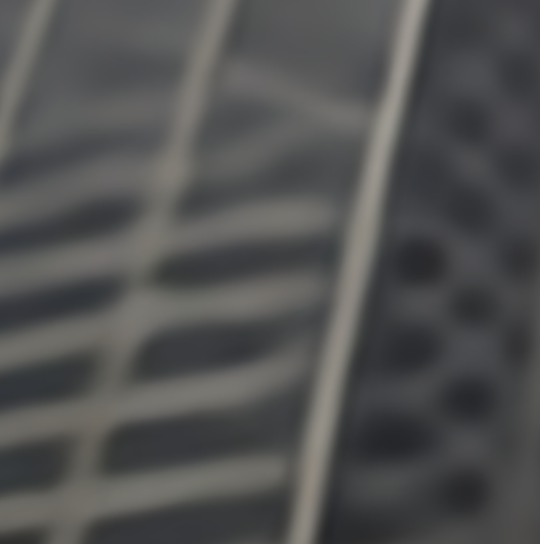} &
   \includegraphics[ width=0.097\linewidth, height=\linewidth, keepaspectratio]{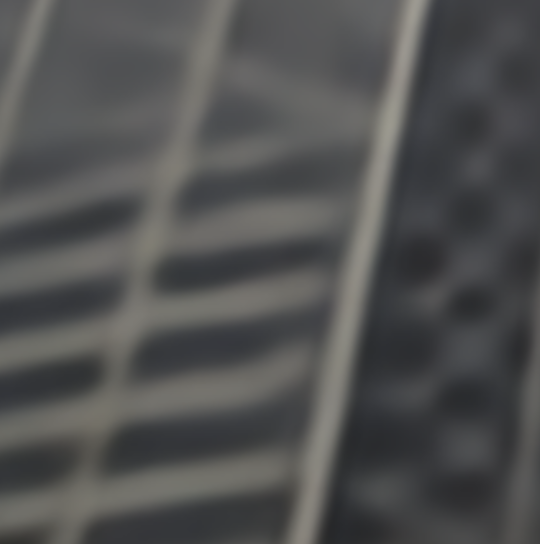} &
   \includegraphics[ width=0.097\linewidth, height=\linewidth, keepaspectratio]{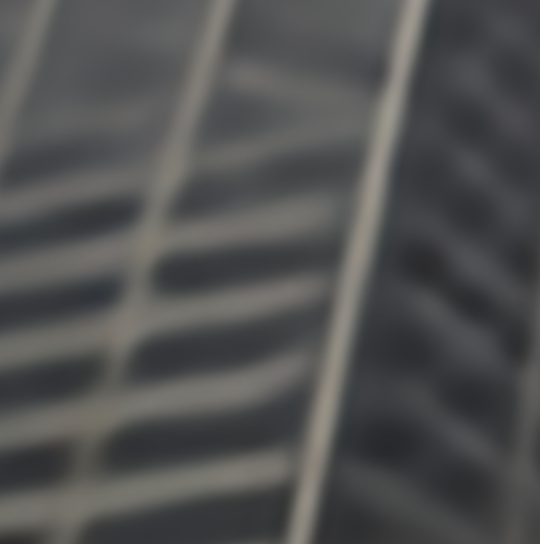} &
   \includegraphics[ width=0.097\linewidth, height=\linewidth, keepaspectratio]{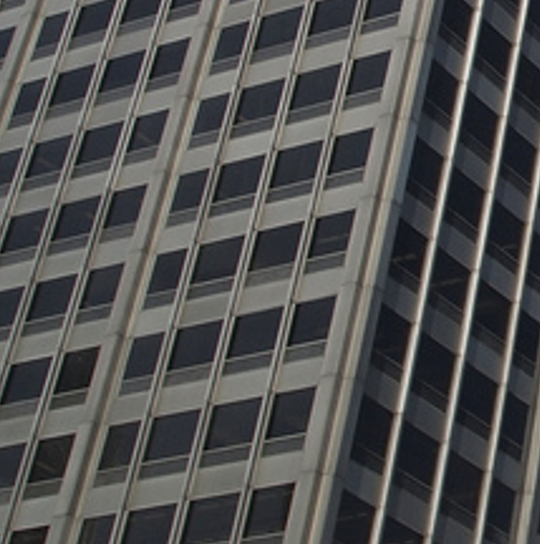}\\
 & LR Image & Bicubic & LIIF~\cite{Authors14} & MoEISR & MoEISR & LTE~\cite{Authors13} & MoEISR& MoEISR & GT \\
 &  &  &  & (LIIF)-s & (LIIF)-b &  & (LTE)-s & (LTE)-b & 
\end{tabular}
}
\caption{\textbf{Qualitative comparison between MoEISR and the backbone models.} RDN~\cite{Authors16} is used as an encoder for all methods.} \label{fig:3}

\vspace{+3EX}
\setlength\tabcolsep{2pt}%
\scriptsize{
\begin{tabular}{cccccc}
   \includegraphics[ width=0.155\textwidth, height=\textwidth, keepaspectratio]{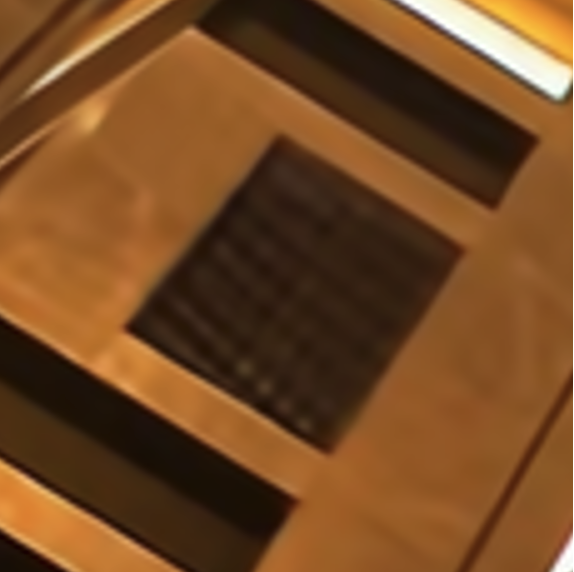} &
   \includegraphics[ width=0.155\textwidth, height=\textwidth, keepaspectratio]{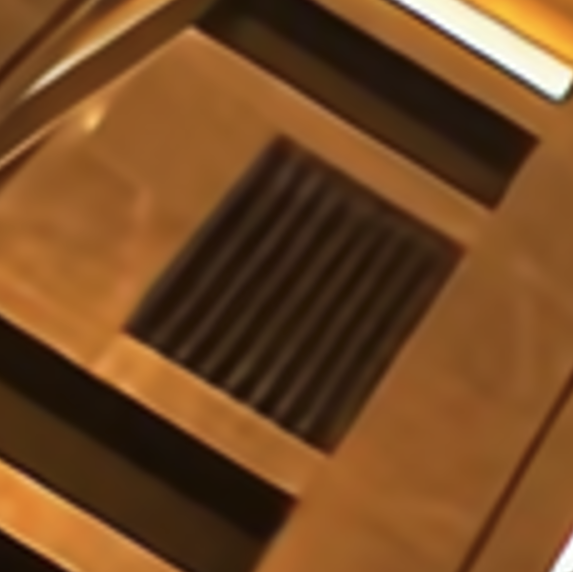} &
   \includegraphics[ width=0.155\textwidth, height=\textwidth, keepaspectratio]{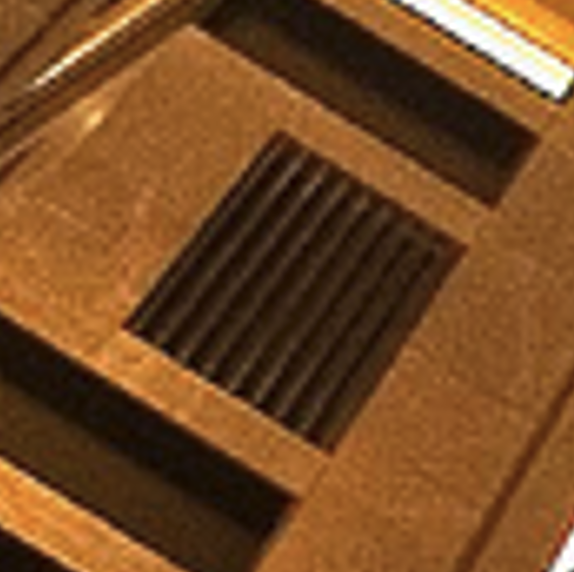} &
   \includegraphics[ width=0.155\textwidth, height=\textwidth, keepaspectratio]{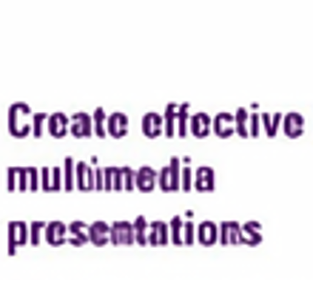} &
   \includegraphics[ width=0.155\textwidth, height=\textwidth, keepaspectratio]{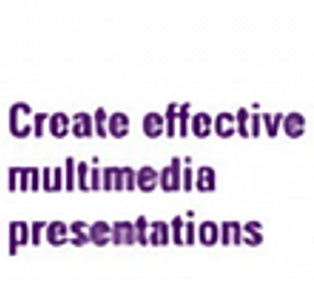} &
   \includegraphics[ width=0.155\textwidth, height=\textwidth, keepaspectratio]{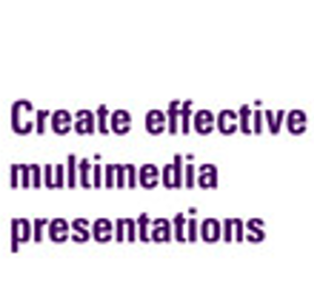}\\
 LTE~\cite{Authors13} & MoEISR(LTE) -b & GT & LIIF~\cite{Authors14} & MoEISR(LIIF) -b & GT \\
  &  &  &  &  
\end{tabular}
}
\vspace{-3EX}
\caption{\textbf{Qualitative comparison between MoEISR and the backbone models.} RDN~\cite{Authors16} is used as an encoder for all methods.}
\label{fig:5}
\end{figure*}

Tab.~\ref{tab:3} provides a detailed description of the FLOPs required to upscale HR images from the Set14 dataset~\cite{Authors20} to various scales. 
As expected, MoEISR, with its adoption of layer-varying experts, consistently yields lower FLOPs for both versions of MoEISR compared to their respective backbone networks. 
Moreover, the FLOPs disparity becomes more pronounced with an increase in the upscaling factor.
The disparity in FLOP reduction between LIIF-based and LTE-based MoEISR models can be attributed to the number of hidden layers employed in the original models. LIIF~\cite{Authors14} utilizes a 5-layer MLP as the decoder, whereas LTE~\cite{Authors13} utilizes a 4-layer MLP as the decoder.
A noteworthy point that deserves attention is that MoEISR -s, which has demonstrated its competitive upscaling ability in Tab.~\ref{tab:1} and Tab.~\ref{tab:2}, requires significantly lower FLOPs than the original model. 
SwinIR-MoEISR(LTE) -s, operating at a scaling factor of 2, requires only 34.9\% of the FLOPs compared to the original SwinIR-LTE~\cite{Authors13} model. 
Furthermore, EDSR-MoEISR(LIIF) -s with a scaling factor of 30 requires even less FLOPs, and it requires only 26.2\% of the FLOPs compared to EDSR-LIIF~\cite{Authors14}.

\textbf{Qualitative Results.} Fig.~\ref{fig:3} provides a comprehensive qualitative comparison on MoEISR and its associated backbone models under in-scale and out-of-scale upscaling factors. 
These models are trained with RDN~\cite{Authors16} as the encoder. 
It is readily apparent that MoEISR outperforms its backbone models in the reconstruction of finer details, regardless of whether the context is in-scale or out-of-scale. 
For instance, the MoEISR(LTE) -b model notably excels in reconstructing square-shaped windows, demonstrating superior performance compared to the LTE~\cite{Authors13} model on both DIV2K~\cite{Authors18} and Urban100~\cite{Authors22} datasets. 
Fig.~\ref{fig:5} visually highlights a noticeable disparity in reconstruction quality between the results of MoEISR and the backbone models.
It is clear that our model demonstrates a high degree of accuracy in reconstructing the window with diagonal patterns, in contrast to the window reconstructed by LTE~\cite{Authors13}.
Furthermore, LIIF~\cite{Authors14} exhibits difficulties in upscaling letters accurately, our model adeptly reconstructs the letters with minimal deformation in shape. The observed discrepancy in the representation of finer details is attributed to the effective utilization of the mapper, which enables our model to capture intricate relationships between individual pixels, thereby enhancing its ability to reconstruct detailed images.

\begin{table*}[t]
    \caption{\textbf{Ablation study on MoEISR and LIIF~\cite{Authors14} with different decoder depths.} PSNR and FLOPs are evaluated with the downscaled Urban100 dataset~\cite{Authors22}. Given that the encoder's FLOPs of all methods are identical, only the decoder's FLOPs are reported. The decoder in MoEISR consists of a router and a set of experts, while LIIF~\cite{Authors14} comprises a single MLP decoder.
    -5layer, -4layer, -3layer and -2layer refers to the decoder depth. {\color{red}Red} and {\color{blue}blue} indicate the best PSNR and least FLOPs, respectively.  }
    \label{tab:4}
    \centering
    \scriptsize{
    \resizebox{\textwidth}{!}{%
    \begin{tabular}{c|cc|cc|cc|cc|cc}
        \multirow{2}{*}{Method} & \multicolumn{6}{c|}{In-scale} & \multicolumn{4}{c}{Out-of-scale}\\
        & \(\times\)2 & GFLOPs & \(\times\)3 & GFLOPs & \(\times\)4 & GFLOPs & \(\times\)6 & GFLOPs & \(\times\)8 & GFLOPs \\ \hhline{===========}
        RDN-MoEISR(LIIF) -b & \color{red}32.93 & 861.84 & \color{red}28.90 & 862.43 & \color{red}26.74 & 864.89 & \color{red}24.24 & 864.26 & \color{red}22.84 & 871.1 \\
        RDN-MoEISR(LIIF) -s & 32.82 & \color{blue}298.75 & 28.80 & \color{blue}296.08 & 26.65 & \color{blue}295.55 & 24.15 & \color{blue}292.90 & 22.76 & \color{blue}294.15 \\ 
        LIIF -5layer~\cite{Authors14} & 32.87 & 1072.33 & 28.82 & 1070.48 & 26.68 & 1071.20 & 24.20 & 1064.99 & 22.79 & 1069.79 \\ 
        LIIF -4layer & 32.83 & 869.14 & 28.80 & 867.64 & 26.64 & 868.22 & 24.15 & 863.19 & 22.75 & 867.08 \\ 
        LIIF -3layer & 32.85 & 665.94 & 28.84 & 664.79 & 26.67 & 665.24 & 24.13 & 661.38 & 22.71 & 664.36 \\ 
        LIIF -2layer & 32.48 & 462.74 & 28.44 & 461.95 & 26.24 & 462.26 & 23.83 & 459.58 & 22.49 & 461.65 \\ 
        
    \end{tabular}%
    }
    }
    \vspace{-1mm}
\end{table*}

\begin{figure}
\setlength\tabcolsep{2pt}%
\footnotesize{
\begin{tabular}{cccccc}
   \includegraphics[ width=0.157\textwidth, height=\textwidth, keepaspectratio]{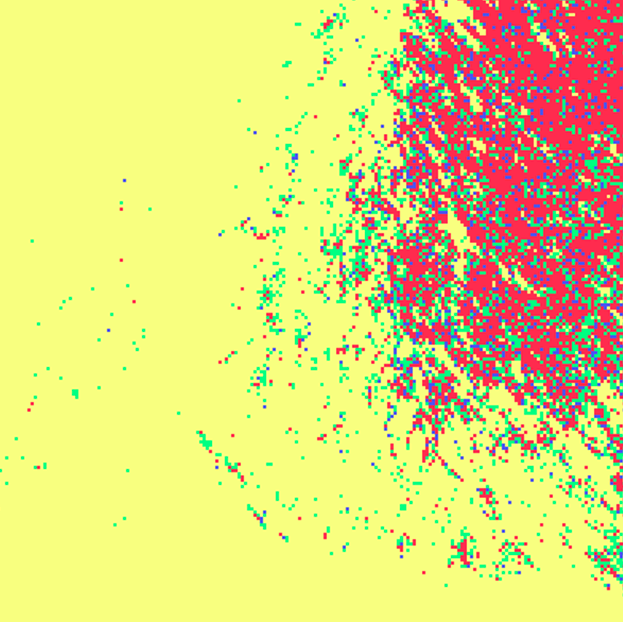} &
   \includegraphics[ width=0.157\textwidth, height=\textwidth, keepaspectratio]{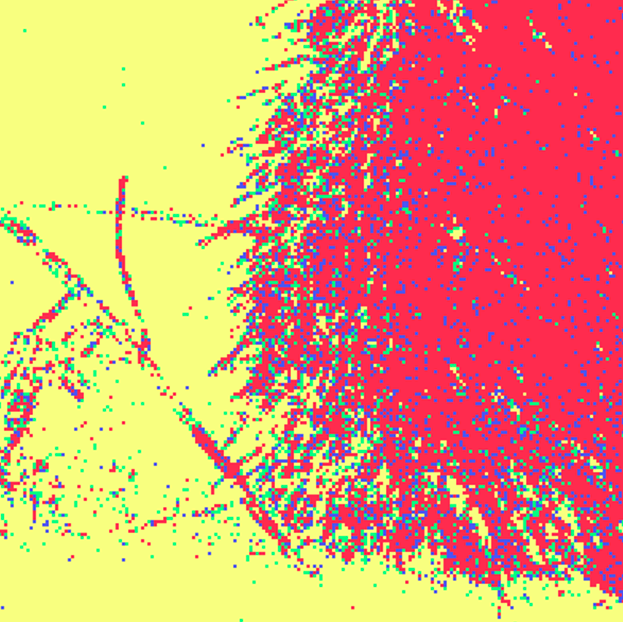} &
   \includegraphics[ width=0.157\textwidth, height=\textwidth, keepaspectratio]{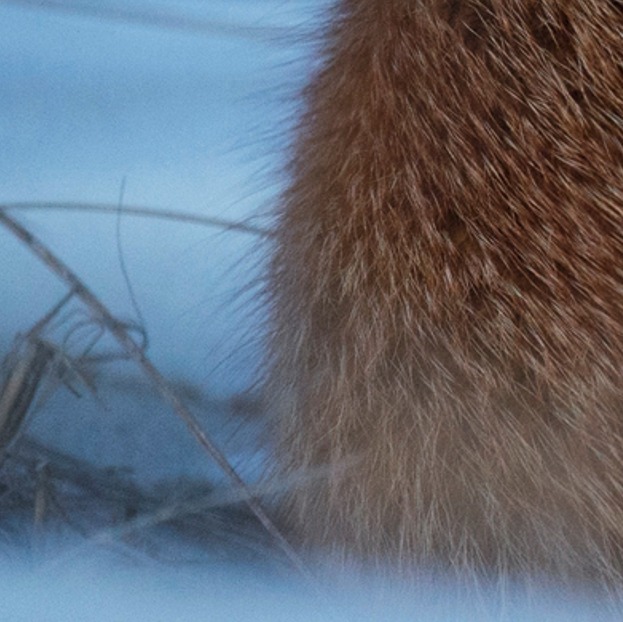} & 
    \includegraphics[ width=0.157\textwidth, height=\textwidth, keepaspectratio]{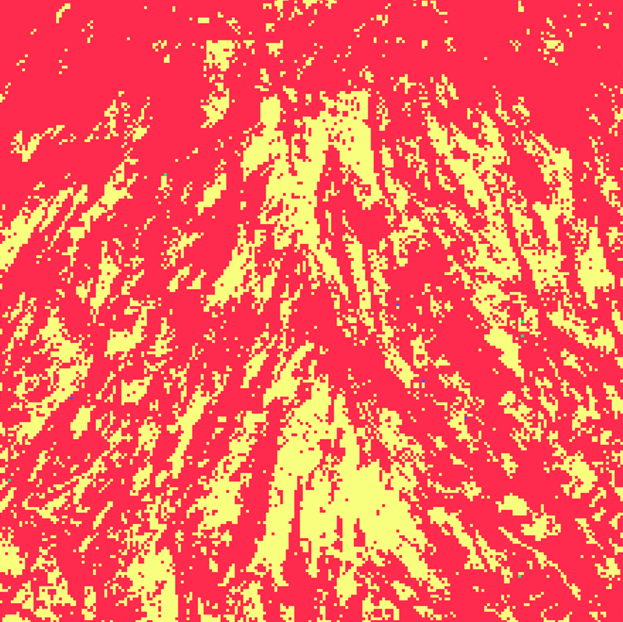} &
   \includegraphics[ width=0.157\textwidth, height=\textwidth, keepaspectratio]{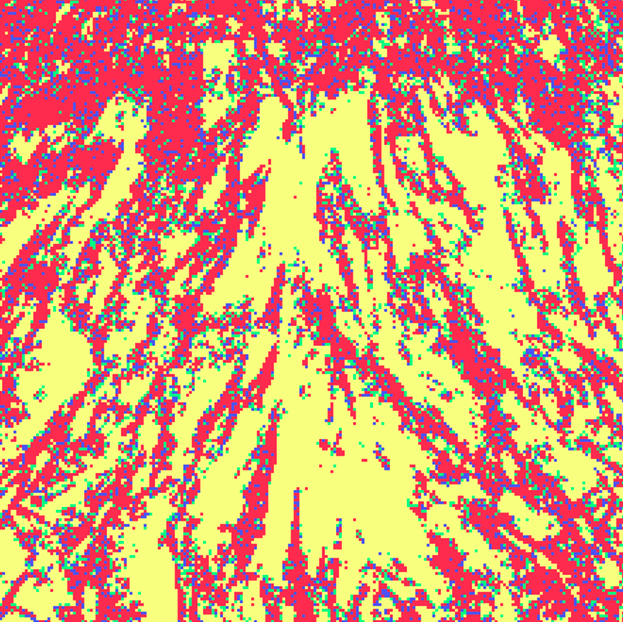} &
   \includegraphics[ width=0.157\textwidth, height=\textwidth, keepaspectratio]{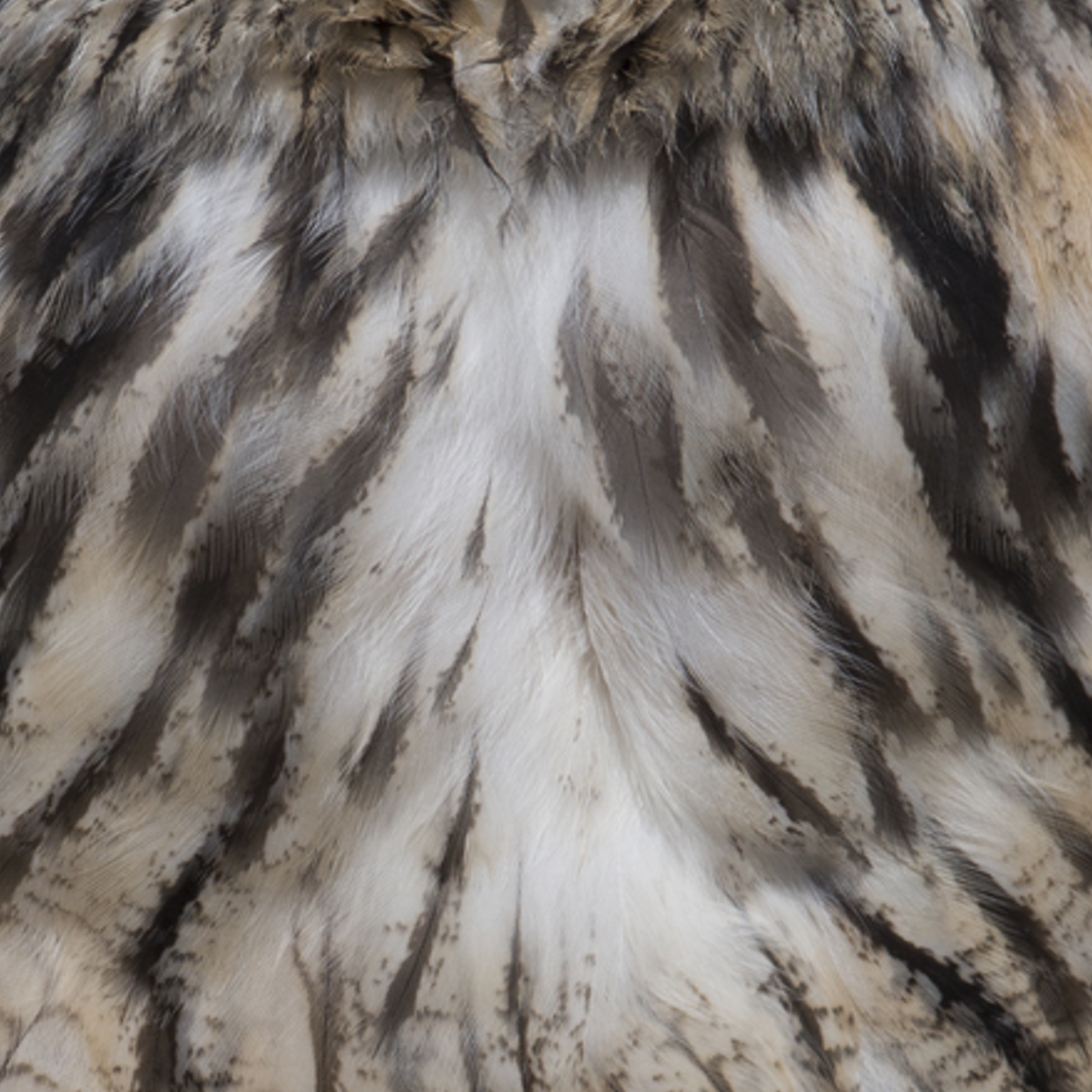}\\
 1-layer mapper & 5-layer mapper & Input image & \(\tau=1\) & \(\tau=5\) & Input image
\end{tabular}
}
\caption{\textbf{MoEISR with different configurations.} Expert map (experts chosen with the highest probability) of 1-layer mapper (leftmost), 5-layer mapper (second-left), corresponding input image (third-left), \(\tau=1\) (third-right), \(\tau=5\) (second-right) and corresponding input image (rightmost). Yellow, green, blue, and red pixels denote different layered experts.}
\label{fig:8}
\vspace{-2EX}
\end{figure}

\begin{table}[t]
    \caption{\textbf{Ablation study on different design choices of MoEISR.} All methods are evaluated on 4\(\times\) bicubic downscaled DIV2K validation dataset~\cite{Authors18}. -m1 refers to MoeISR with a 1-layered mapper, $\tau$3 refers to MoEISR with temperature hyperparameter $\tau$5 refers to MoEISR with temperature hyperparameter $\tau=5$.}
    \label{tab:5}
    \centering
    \footnotesize{
    \begin{tabular}{c|ccc}
        Method & PSNR & SSIM & FLOPs  \\ \hhline{====}
        MoEISR(LIIF) -b & 29.2909 & 0.8273 &  6.84TFLOPs\\
        MoEISR(LIIF) -b -$\tau$3 & 29.2099 & 0.8264 & 6.43TFLOPs\\
        MoEISR(LIIF) -b -$\tau$5 & 29.1811 & 0.8260 &  6.39TFLOPs\\
        MoEISR(LIIF) -b -m1 & 29.1685 & 0.8251 & 5.61TFLOPs\\
    \end{tabular}%
    }
\vspace{-2EX}
\end{table}

\subsection{Ablation Study}
In this section, we perform a comprehensive ablation study of each component within MoEISR and evaluate its impact on the overall performance.

\textbf{Fixed Decoder.} To assess the effectiveness of MoEISR, we compare our models to LIIF~\cite{Authors14} with various decoder configurations. RDN~\cite{Authors16} is used as an encoder for all models and is evaluated with the Urban100 dataset~\cite{Authors22}. Tab.~\ref{tab:4} describes the differences in PSNR and FLOPs across varying model architectures and the notations -5layer, -4layer, -3layer, and -2layer denote the respective number of convolutional layers within the decoder. The acquired outcomes undeniably demonstrate the efficacy of MoEISR. Specifically, MoEISR(LIIF) -b consistently achieves the highest PSNR while maintaining a comparable level of FLOPs as that of LIIF -4layer, which yields about 0.1db lower PSNR than MoEISR(LIIF) -b. Moreover, it is worth noting that MoEISR(LIIF) -s consistently demands the least amount of FLOPs while still achieving competitive PSNR values, similar to those of LIIF -4layer, which necessitates significantly larger FLOPs.

\textbf{Mapper Depth.} The mapper module, which is responsible for determining the appropriate expert for restoring each output pixel, consists of five convolutional layers. 
Inspired by APE~\cite{authors30}, which employs a single-layered regressor, we explore a modified version of MoEISR. This variant incorporates a notably lighter mapper module, comprising only a single convolutional layer.
Tab.~\ref{tab:5} provides a clear representation of the performance of MoEISR with varying mapper depths. It is worth noting that the utilization of a 1-layered mapper module results in a reduction of FLOPs, but there is also a corresponding decrease in both PSNR and SSIM.
Fig.~\ref{fig:8} visually depicts the expert maps generated by the mapper modules with varying depths. It is evident that the 5-layered mapper produces a more intricate expert map compared to the 1-layered mapper. For instance, 5-layered mapper adeptly captures the fine details of the grass adjacent to the fur, whereas 1-layered mapper struggles to assign the appropriate decoders to the area.
Nevertheless, since our mapper is executed only once like the encoder, the use of the 5-layer mapper does not significantly impact on the overall FLOPs during the \(4 \times\) high-resolution image reconstruction process.

\textbf{Temperature Hyperparameter.} In our approach, we employ Gumbel-softmax~\cite{Authors26, authors49} with temperature hyperparameter \(\tau=1\). 
This temperature hyperparameter plays a key role in the assignment of experts to the output pixels and has a substantial impact on the efficacy of MoEISR.
Fig.~\ref{fig:8} shows different expert maps generated with different temperature hyperparameters. 
In MoEISR, we do not force the mapper to assign experts equally, therefore achieving a balanced trade-off between performance and efficiency. This behavior is accomplished by configuring the hyperparameter \(\tau\) to a value of 1, thus intensifying the probability discrepancies among different experts.
As described in Tab.~\ref{tab:5}, when the hyperparameter \(\tau\) is set to 3, our Mapper yields a lower PSNR and SSIM values with reduced FLOPs, and this effect becomes more pronounced as the value of \(\tau\) increases.

\textbf{Controllable Mapper.} We conduct additional experiments to explore the extent to which we can control the trade-off between the speed and quality of the model. In Eq.~\eqref{eq:6}, hyperparameter $w_j$ is multiplied to control the assignment of experts to the output pixels. 
By simply altering the values of \(w_j\) during the training phase, we can effectively determine the frequency of assignments for each expert.
The visualization of expert map with varying $w_j$ can be found in the supplementary materials.
When $w_4$ is set to 3.7, the mapper module allocates the 4-th expert (5-layered expert) less frequently, 
while setting \(w_1\) to 3.7 leads to reduced usage of the 1-st expert (2-layered expert). 
This control over expert allocation allows us to balance the computational load and restoration quality effectively.


\section{Conclusion}

In this paper, we present MoEISR, a novel approach that achieves the dual goal of significantly reducing computational requirements while maintaining competitive SR image quality. 
One of the key advantages of our method is its versatility, as it can be smoothly integrated into any INR-based arbitrary-scale SR framework. 
MoEISR consists of three core components: an encoder, a mapper, and a set of experts. 
The encoder is responsible for extracting INR from the input image, while the mapper generates an expert map that assigns the most suitable expert to each output pixel regarding its reconstruction difficulty. 
The central concept behind MoEISR is the use of a mapper module in conjunction with a set of experts with varying depths, allowing each pixel to be reconstructed with the most suitable expert. By using multiple experts, as opposed to a single computationally-heavy decoder, each expert specializes in reconstructing different regions within the image, ultimately resulting in high-quality SR images with a reduced computational load. Extensive experiments presented in the paper demonstrate the ability of MoEISR to improve the quality of existing INR-based arbitrary-scale SR networks, while simultaneously significantly reducing computational requirements. 

\clearpage

\bibliographystyle{splncs04}
\bibliography{NeurIPS24/neurips_2024}

\end{document}